\documentclass[sigconf]{acmart} 
\usepackage[english]{babel}
\usepackage{blindtext}
\usepackage{cuted}
\usepackage{lineno}
\usepackage{cmbright}
\usepackage{fancyhdr}
\renewcommand\footnotetextcopyrightpermission[1]{} 
\setcopyright{none}

\acmConference[]{}
\acmYear{}
\copyrightyear{}

\title{Solar-powered shape-changing origami microfliers}

\begin{document}
\renewcommand{\shortauthors}{X.et al.}

\author{\large Kyle Johnson$^{1\dagger}$, Vicente Arroyos$^{1\dagger}$, Am\'elie Ferran$^{2,3}$, \\ Tilboon Elberier$^{4}$, Raul Villanueva$^{4}$, Dennis Yin$^{4}$, \\Alberto Aliseda$^{2}$, Sawyer Fuller$^{1,2}$, Vikram Iyer$^{1\ast},$ 
Shyamnath Gollakota$^{1\ast}$\vspace{0.1in}\\
$^{1}$ Paul G. Allen School of Computer Science and Engineering, University of Washington\\
$^{2}$ Department of Mechanical Engineering, University of Washington\\
$^{3}$ LEGI Laboratory, Universit\'e Grenoble Alpes \\
$^{4}$ Department of Electrical and Computer Engineering, University of Washington\\
$^\dagger$ Equal contributing  first authors, 
$^\ast$Corresponding authors}

\thispagestyle{fancy}
\maketitle
\thispagestyle{fancy} 
\pagestyle{plain} 

\begin{strip}
\begin{large}
\centering\textbf{ABSTRACT} \\
\end{large}
\vskip 0.05in 
\raggedright Using wind to disperse microfliers that fall like seeds and leaves can help automate large-scale sensor deployments. Here, we present battery-free microfliers that can change shape in mid-air to vary their dispersal distance. We design origami microfliers  using bi-stable leaf-out structures and uncover an important property: a simple change in the shape of these origami structures  causes two dramatically different falling behaviors. When unfolded and flat, the microfliers exhibit a tumbling behavior  that increases lateral displacement in the wind. When folded inward,  their orientation is stabilized, resulting in a downward descent that is less influenced by wind. To electronically transition between these two  shapes, we designed a low-power  electromagnetic actuator that produces peak forces of up to 200~millinewtons within 25 milli- seconds while powered by solar cells.  We fabricated a circuit directly on the folded origami structure  that includes a  programmable microcontroller,  Bluetooth radio, solar power harvesting circuit, a pressure sensor  to estimate altitude and a temperature sensor. Outdoor evaluations show that our 414~milligram origami microfliers are able to electronically  change their shape mid-air, travel up to 98~meters in a light breeze, and wirelessly transmit  data via Bluetooth up to 60~meters away, using only  power collected from the sun.
\end{strip}

\section*{Introduction}
Many plants passively disperse biological material in the wind such as seeds and leaves~\cite{dandelion-flight,somesciencepaper,dandelion-morphing}. This ability to   disperse in the wind without  active propulsion is useful for designing {wind-dispersed microfliers~\cite{dandelion-sensors,microflier}. Equipped with sensors, such  microfliers could  automate the deployment of large-scale wireless  sensor networks for environmental monitoring~\cite{microflier}. These designs are much smaller and lighter than drones~\cite{liquid-mav, mav-forceful, mav-swarm}, however they lack onboard actuation and thus do not have in-air control over varying their descent behavior or dispersal distance. {In this work, we  engineer miniaturized, battery-free, programmable microfliers that can both disperse in the wind and vary their dispersal distance through electronic actuation. The actuation can be triggered  either by  an onboard sensor or through wireless communication.}

Achieving effective wind dispersal requires minimizing the mass of the microfliers to achieve low terminal velocities for maximum flight time. Introducing actuation and control however adds the mass  of the actuation mechanism, requires onboard sensing and computation for control, as well as the ability to power  these components.

Prior designs  use  fixed wing gliders~\cite{CICADA} and spinning seed-inspired designs~\cite{samara,samara-scirob}  to alter their descent behavior; however, these designs use large, power-consuming motors and servos and required heavy batteries. As a result, they are orders of magnitude larger in size and weight than our sub-gram miniaturized microfliers.

{We present battery-free microfliers that are able to  electronically change their shape in mid-air to alter their falling behavior and vary their dispersal distance} (Fig~\ref{fig:fig1}). These solar-powered miniaturized devices can be dropped from small commercial drones. They spread  outward in the direction of the wind in their tumbling state. {Upon reaching a programmable altitude, triggering a timer, or receiving a wireless trigger signal, these devices can use harvested power to transition to a stable state in mid-air, in which they descend with decreased lateral dispersal.} Upon landing, the devices continue to harvest energy to power onboard environmental sensors and a Bluetooth radio to wirelessly  transmit sensor data.

\begin{figure*}[t]
\vskip -0.3in
\centerline{\includegraphics[width=1.\textwidth]{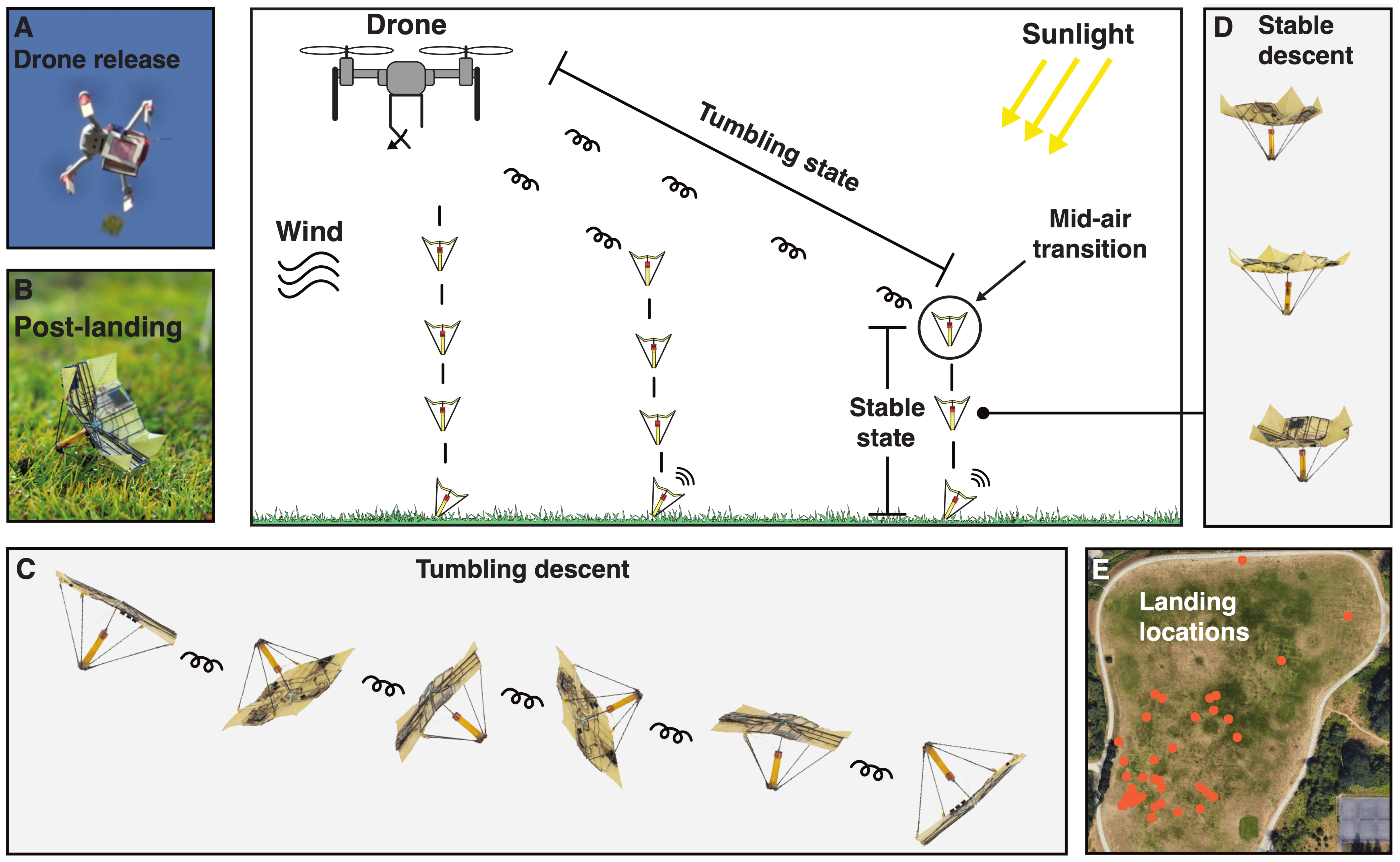}}
\vskip -0.15in
\caption{\textbf{Solar-powered shape-changing origami microfliers.} Conceptual diagram show origami microfliers changing their shape in mid-air to alter their falling behavior and achieve varied sensor dispersal distances. The microfliers are released from a drone (\textbf{A}) and can wirelessly transmit sensor measurements using solar power after landing (\textbf{B}). They begin falling in a flat, tumbling state (\textbf{C}) with greater lateral dispersal in the wind. The microfliers use harvested solar power to electronically transition in mid-air after a programmable time or altitude using their onboard actuators into a folded state (\textbf{D}) with stable descent and lower wind dispersal. This enables individual microfliers to achieve varying dispersal distances as shown in a aerial view (\textbf{E}).}
\label{fig:fig1}
\end{figure*} 

Given the limited and intermittent nature of solar harvesting, instead of continuous actuation, our microflier design uses leaf-out origami based on the Miura-ori building block to produce bistable structures~\cite{science2005,leafout1,leafout2}.
These structures maintain their configuration in either of their two states without any active energy consumption.  

Our work  reveals an important property of leaf-out origami: these structures  have dramatically different falling behaviors in their two states. Fig~\ref{fig:fig2}A,B show comparisons of the flat ``tumbling'' state in which the structure falls chaotically, and the folded state in which it exhibits a stable descent. A fluid analysis demonstrating the airflow around the microflier in each of its two states is shown in Fig~S1. Movies S1,2 demonstrate that a small inward fold can emulate two distinct descending behaviors of two  different leaf shapes, shown for comparison. This difference in behavior substantially changes their response to lateral wind gusts (Movie S3), {which can be used to vary their wind dispersal distance}. 

\begin{figure*}[t]
\vskip -0.3in
\centerline{ \includegraphics[width=1.\textwidth]{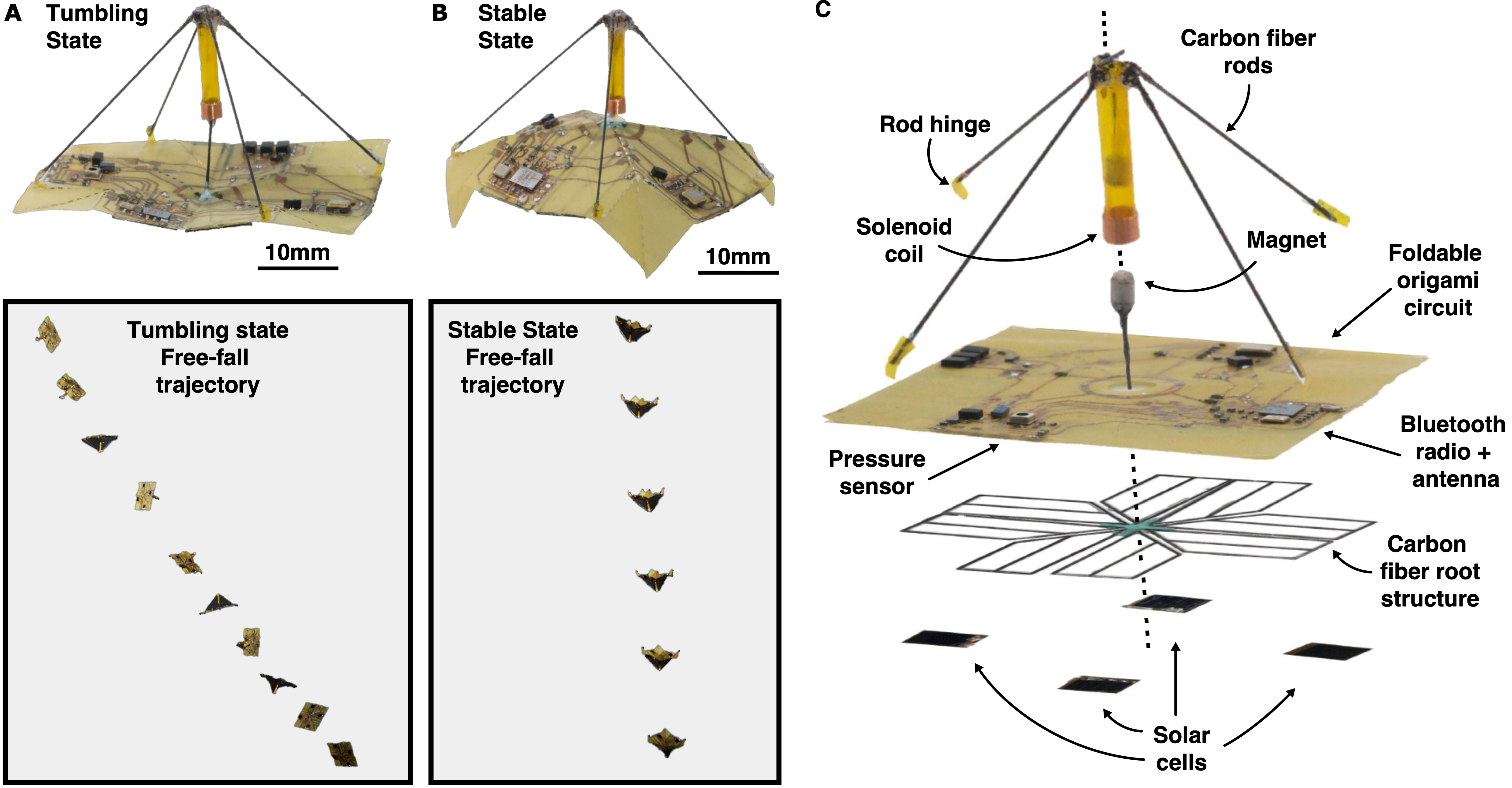}}
\vskip -0.15in
\caption{\textbf{Tumbling and stable states.} \textbf{A}) Origami microflier in its tumbling state. The free fall trajectory shows sampled video frames of the flipping and lateral motion. \textbf{B}) Origami microflier in its stable descent state. The free fall trajectory shows stable descent and minimal lateral motion. \textbf{C}) Disassembled  view of our origami microflier showing the major components.}
\label{fig:fig2}
\end{figure*} 

Creating a solar-powered origami microflier is challenging for three  reasons.  First, the structure should not transition until actively triggered. As the microflier tumbles, it not only encounters the force of the wind, but also gravity acting on the mass of the payload. This imposes a trade-off between the origami design and the actuator. The structure must be stiff enough to prevent false transitions, but doing so also increases the force the actuator has to deliver, which in turn requires larger components and higher power consumption. Second, transitioning the solar-powered microflier in mid-air imposes the strict requirement that it must be completely untethered from any power source or actuation stimulus, be electronically controllable by the device itself, and produce a rapid response to transition before the device falls to the ground. Although origami systems  that use external magnetic fields, shape memory alloys, electrothermal polymers, motors, peizoelectric actuators or electrostatic actuators have been proposed in the literature~\cite{bergbreiter-review,ingestible-origami, miyashita2015untethered,magnetic-programmable-matter,sma-origami-2010, robogami,printed-electrothermal,thread-origami, belke2017mori,robofly,electrostatic-origami,tribot-nature}, none of them meet our size, weight, power, and rapid response requirements.  Third,  this system must  rapidly charge a lightweight energy storage element such as a capacitor up to the voltage required by the actuator, and discharge a pulse of energy sufficient for transitioning.  Moreover, many microcontrollers are designed for low voltage operation to minimize power consumption and cannot tolerate the higher  voltages that may be required for actuation. This requires creating separate power regulation circuits for both parts of the system and a strategy to dynamically switch between them to multiplex a single lightweight solar array.

In this article, we demonstrate that it is possible to address these challenges and build solar-powered origami microfliers that can electronically change their shape in mid-air for wind dispersal of wireless sensors. We make the following key contributions: First, we design  origami microfliers and demonstrate that small changes in their shape can dramatically change their falling behavior from chaotic tumbling to a stable descent. Further, we observe that the microfliers are more responsive to lateral wind gusts in their tumbling state achieving as much as 3-fold greater dispersal distance than in their stable state.  Second, we  combine our origami with a low-power, bi-stable electromagnetic actuation mechanism compatible with solar power harvesting. {This allows our origami microflier to transition between origami states in mid-air, completely untethered. Our onboard actuator produces peak forces of up to 250~mN within 25~ms. This enables the design of robust origami microfliers that can operate without false transition in wind speeds upwards of 5~m/s.} Third, we   design a solar energy harvesting circuit that   can both cold-start from zero charge at sunrise and harvest enough power to transition the structure in mid-air. Additionally, an onboard microcontroller, radio, and pressure sensor enable multiple modes of operation for triggering a transition based on time delays, altitude readings, or wireless commands. After landing, the onboard temperature and pressure sensor wirelessly transmit data to a remote Bluetooth receiver at distances of 60 m. Finally, we perform real-world deployments by dropping our sensors from drones at altitudes of  40~m and demonstrate dispersal up to distances of 98~m in a light breeze.  {We further show that our 414~mg device can harvest enough energy to transition in mid-air. Additionally, the devices can receive trigger signals via Bluetooth and transmit real-time sensor data as they fall for taking sensor measurements at different altitudes while landing with their solar cells facing up 87\% of the time.}
\section*{Results}

\begin{figure*}[t]
\vskip -0.3in
\centerline{ \includegraphics[width=1.0\textwidth]{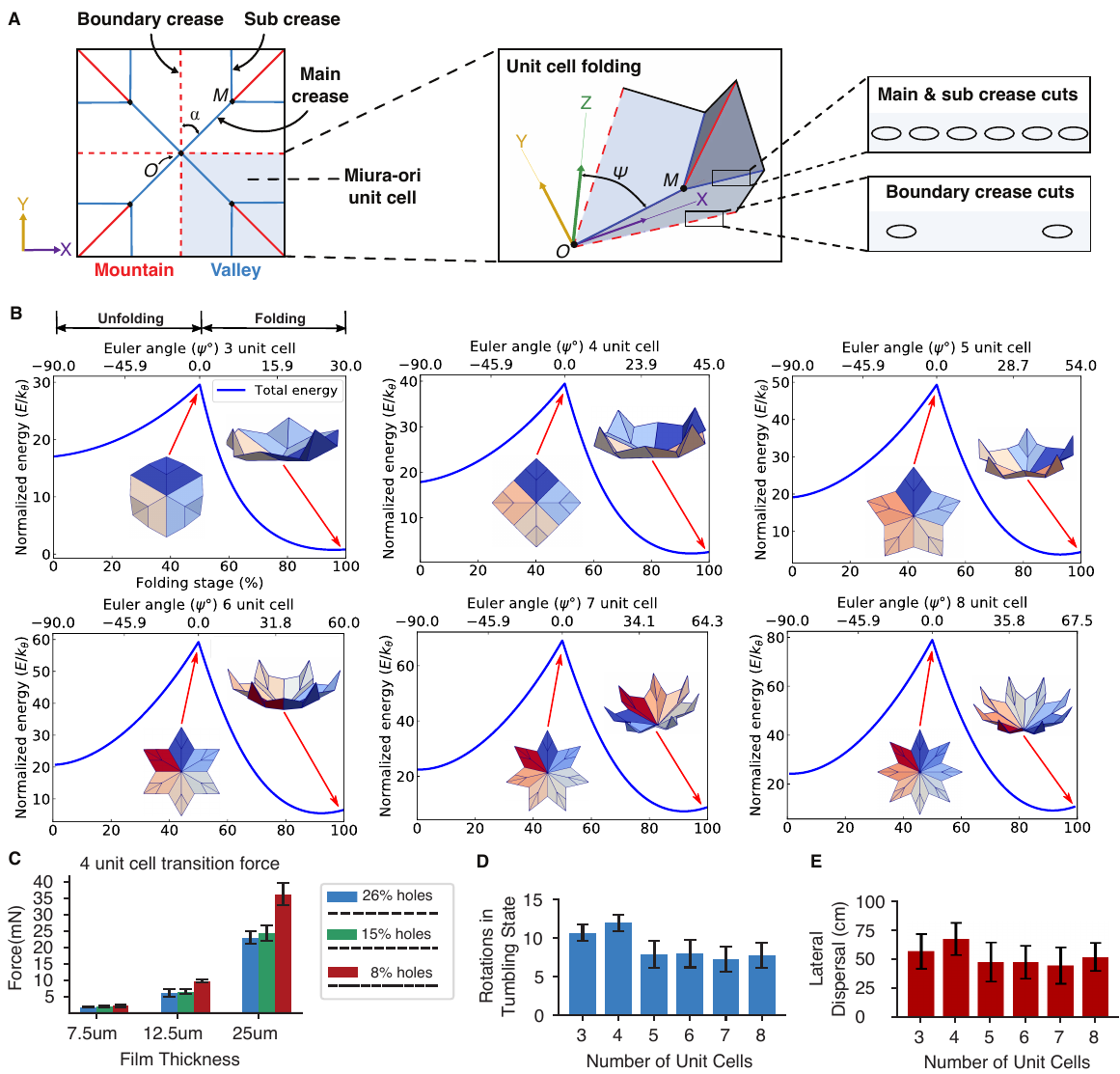}}
\vskip -0.15in
\caption{\textbf{Bi-stable origami structure.} \textbf{A}) Diagram showing key features of the Miura-ori fold and tessellation to create structures with multiple unit cells. Sample cut patterns that add holes along creases to tune the stiffness of origami folds are shown to the right. \textbf{B}) Kinematics simulation showing energy required for designs with different numbers of unit cells to transition between states. \textbf{C}) Force required to transition a four unit cell origami structure on different film thicknesses and various cut patterns, without electronics or actuation system. \textbf{D}) Number of flips observed when dropping designs with various numbers of unit cells in the tumbling state from 2 m ($N=10$, $\pm\sigma$). \textbf{E}) Lateral displacement when dropped from height of 0.5~m in the tumbling state ($N\geq 10$, $\pm\sigma$).}
\label{fig:origami}
\end{figure*} 

\noindent{\bf Bi-stable origami structure.} 
Our objectives were to design miniaturized  (sub-gram), wind-dispersed microfliers that have the ability to change their falling behavior to thereby vary their dispersal distance. One method to achieve this is to use active propulsion such as spinning rotors that can produce lift;  this, however, requires substantial energy and is challenging to achieve using solar-harvesting~\cite{solar-crazyflie}. A lower power alternative would instead be to leverage interactions with the air flowing around the microflier as it falls. In particular, the terminal velocity of a falling object is affected by its projected area and structure dependent drag coefficient~\cite{dandelion-flight}. Changing the shape of the falling microflier therefore presents an alternative means of changing its descent behavior.

To create a shape change with minimal energy, we used  bi-stable origami  structures that can alternate between two different folded shapes. Specifically, we leveraged the bio-mimetic Miura folds that occur in leaves~\cite{tree-leaves}. The Miura-ori pattern can be tessellated as a building block for high expansion ratio structures that produce substantial shape changes~\cite{tree-leaves, deploy-origami} and achieve bi-stable structures~\cite{DeF-leaf-like, leafout2}. Additionally, the Miura-ori pattern is a form of rigid origami, meaning the faces of the structure will not contort during folding and deformations only occur along defined crease lines. This provides two key engineering advantages. First, energy is concentrated along the crease lines and not expended, elastically deforming the faces of the structure. This minimizes the energy required to transition the structure from one state to another. Second, the lack of deformation on the faces provides stable areas in which to attach electronic components like solar cells.

We designed our origami microfliers by combining multiple Miura-ori unit cells as shown in Fig~\ref{fig:origami}A. An individual unit cell is shown shaded gray and is composed of three distinct types of crease lines: boundary, main, and sub creases. The structure was  tessellated by connecting adjacent boundary crease lines to other unit cells, with all the unit cells' main and boundary crease lines meeting at the origin point $O$. Each crease is represented as either a mountain fold (protruding out of the page) or a valley fold (going into the page). 

{Following the rigid origami model, we assumed all creases act like hinges between rigid and non-bendable panels (faces). Each unit cell is characterized by a central angle ($\alpha=\pi/n_{cells}$) and main crease line length  ($L=\overline{OM}$) parameter. The crease stiffness of our fabricated prototypes was tuned by using different materials, varying material thickness, and by introducing cut patterns along the crease lines to decrease stiffness, shown to the right in Fig~\ref{fig:origami}A. As the leaf-out is folded from one state to the other, its folding stage is characterized by $\psi$, representing the angle between the main crease line and the Z-axis extending through the center of the structure. Intuitively, this angle represents how much the main crease is folded inward.}

To understand the energy requirements for changing the origami shape, we looked  to prior works on leaf-out origami simulations to model the kinematics of the structure in different configurations \cite{leafout1, leafout2}. It has been shown that we can analyze different configurations of the leaf-out structure using the rigid origami simulation technique, where we assumed that the crease line folds are represented as torsional springs, each Miura-ori cell exhibits a single degree-of-freedom (DOF) folding motion, and transformations are assumed to be uniform \cite{leafout1, leafout2, waterbomb, rig-orig2}. A uniform transformation has all $n_{cells}$ with equal $\psi$ values throughout each folding stage, and all mountain and valley crease assignments maintained throughout each folding stage. Using the assumptions stated above, it has been shown that we can estimate the peak energy required to transition the structure by conducting an energy analysis based on the kinematics of the leaf-out structure \cite{leafout1,leafout2}. Fig~\ref{fig:origami}B shows the energy required to transition the structure between states  for a variety of leaf-out configurations with different $n_{cells}$. This value represents the sum of the energy required for each fold normalized by their spring constant $\kappa_\theta$ (see Supplementary Materials for details). We assumed this constant is the same across folds as they are made from a uniform material. The normalized energy plots show that as $\psi$ approaches  0\textdegree, the flat state, the energy required to fold the structure to that angle increases rapidly. Beyond this peak, the structure transitions to its other state and the energy rapidly decreases. Movie S4 shows how the structure folds and transitions between states. These results show that increasing the number of unit cells increases the peak energy required to transition, another parameter that can be tuned to match the ideal configuration for a given application. In this work, we optimized the $n_{cells}$ of our leaf-out design to maximize lateral displacement for enabling dispersal in the wind over a wide area. We then tuned the material thickness and crease cut pattern to ensure that the chosen leaf-out configuration can be transitioned given the fixed energy generated from our actuator.

Next, we evaluated how lightweight origami structures fabricated on thin films fall in each of their states. To characterize this behavior, we performed empirical measurements due to the complexities of accurately modelling the fluid–structure interactions for thin, deformable, freely falling objects. We used  laser micromachining to cut out each of the simulated structures from polyimide films (see Methods for details). To tune the energy required to transition, we experimented with altering the film thickness, as well as by varying the cut patterns on the boundary and sub crease lines to vary their stiffness and therefore the force required to transition. 

Specifically, we evaluated  the force required to transition four unit cell origami structures made with different three different polyimide film thicknesses and boundary crease cut patterns. We note that these experiments were performed on the bare folded film. We measured the force by placing the origami folded polyimide sheets onto a precision weight scale (Sartorius QUINTIX125D-1S) and position a screw driven micromanipulator holding a flat carbon fiber plate affixed to a rod above the center of the origami. We slowly moved the carbon fiber rod and plated down to apply pressure to the origami and recorded the reading on the scale at which it transitions. We then converted the measured mass to a force in Newtons and plotted the results in Fig~\ref{fig:origami}C. {The legend indicates the percentage of cut length, or the percentage of holes along the crease line. A higher percentage indicates more holes along the crease and therefore lower stiffness. We performed  additional characterization of how these values change when the structure has payload such as circuit components on the faces in Fig~S2. These experiments showed a range of forces that can be achieved by modulating the material and cut parameters.} 

These results demonstrated the ability to achieve a variety of values to suit potential actuation strategies. During these tests however we observed that off axis forces and airflow caused these thin structures to warp and violate the rigid origami design principle. In order to address this, we again looked to real leaves for inspiration and noted that they have  rigid veins that define their structure. We emulated this by creating patterns of carbon fiber and PET as shown in Fig~\ref{fig:fig2}C. We attached these carbon fiber root structures to the faces of the origami to create multilayered prototypes (see Methods and Movie S4). We further found that this increases the force required to transition. For example, a 12.5~{\textmu}m thick leaf-out structure with 26\% holes along the main and sub creases goes from requiring 6.5~mN to transition without the root structure to 34~mN with the root structure. 

{We evaluated the microflier's resistance to false transitions in high winds by suspending the structure from a thin Kevlar thread above a fan, and exposing  the origami face to wind speeds up to 5~m/s. Movie S5 showed that this wind speed was strong enough to break the glue joint holding the flier in place, after which it goes into the air and lands on the ground. We note that even after experiencing this force, the structure did  not falsely transition, and these results were consistent across 10 experiments averaging 7~s in duration. We further note that by tuning the material thickness and crease cuts we could make the structure even more resilient to false transitions.}

Next, we performed a series of drop tests with our origami microfliers and observed their behavior during free fall. We fabricated an array of origami designs with 3-8 unit cells on 12.5~{\textmu}m thick Kapton sheets, and performed drop tests (N=10 trials) for each origami design in its two folded states from a height of 2~m. We recorded videos of their descent, and observed that our designs reach their terminal velocity from this height. 

These experiments highlighted a key difference in the behavior of the origami in its two states as shown in Movies S1, S2 and S3. When the structure was flat it quickly began  tumbling about an axis, whereas in its closed state it exhibited a stable descent. We further observed  that the tumbling state was more affected by wind gusts, giving it the potential to travel longer lateral distances during wind dispersal. {The results of our drop tests comparing different origami designs in their flat and tumbling state are shown in Fig~\ref{fig:origami}D,E. We observed that the four unit cell design produced a greater number of rotations as it descends. This is correlated with the observation that the four unit cell design achieved  greater lateral displacement from the drop location, suggesting that the momentum from this rotation causes it to move further outward from the drop location. Additionally, as shown in Movie S3, flipping contributes to greater lateral displacement in wind gusts}. We also performed  experiments analyzing falling behavior with payload and different weight distributions shown in Fig~S3 and provide additional details in supplementary methods.

{To understand the cause of this behavior, we performed particle image velocimetry (PIV) measurements to characterize the flow around and in the wake of the flier, using the setup described in Fig~S1A and Supplementary methods (see PIV analysis). In its flat tumbling state, the flier presented sharp edges to the incoming air flow as it falls downward. We observed  that the flat, tumbling configuration has a wider, highly asymmetric instantaneous wake, with higher frequency vortex shedding which results in large aerodynamic torque with respect to the flier’s center of mass. Specifically, Fig~S1B shows the magnitude of the aerodynamic torque, represented by the location of the center of pressure with respect to the center of the flier is substantially higher in the flat, tumbling state than in the stable state. The location of the center of pressure moved over a range of 10 mm at characteristic frequencies on the order of 10~Hz causing the flier to tilt and begin tumbling. This represents the main mechanism causing the instability. Once in unstable fall, there was no aerodynamic torque to return the flier to stable fall, perpendicular to its plane. In contrast, the folded stable state presents beveled corners to the flow, resulting in narrower wake, as shown in the time-averaged flow velocity contours in  Fig~S1C. In this state, the wake vortices formed closer to the center of the flier, resulting in a more symmetric wake and lower aerodynamic torque with respect to its center of mass. Thus, the flier in the folded state is more stable: less prone to tilting its plane from falling perpendicular-to-gravity to aligned-with-gravity. Experimental methods and analysis are discussed extensively in Supplementary Methods (see PIV analysis).}

\vskip 0.05in\noindent{\bf Solar-powered actuator.}
Creating an origami microflier that can transition in mid-air between tumbling and stable states requires careful co-design between the origami structure, actuator, and power harvesting circuit which raises multiple design challenges. First, the structure must be robust to false transitions when it encounters the force of the wind and gravity acting on the mass of the payload. This presents a trade-off between the origami design and the actuator. The structure must be stiff enough to prevent false transitions, but doing so also increases the force the actuator has to deliver, which in turn requires larger components and higher power consumption. Second, the actuation mechanism itself must be compatible with the complex geometry of the origami structure and tolerate folding. As shown in Fig~\ref{fig:origami}A and Movie S4, when the origami structure transitions between states, the center point and creases move up and down along the Z-axis whereas the borders of the structure contract inward along the X and Y axes. This makes it challenging to mount a  rigid actuator on the structure at a fixed mounting point that is compliant with the origami folds. 
Third, the actuation mechanism must produce a rapid response to transition the device before falling to the ground. We observed that when dropped from an altitude of 40~m, our microfliers were airborne for approximately 15~s. This means that our actuator must be able to transition much faster than this to achieve our target of programmable transitions at different heights.
Fourth, transitioning the structure in mid-air imposes a  strict requirement that the microflier must be completely untethered from any power source or actuation stimulus and be electronically controllable by the device. To achieve this in a lightweight form factor we used solar power harvesting; this, however,  adds  constraints on both the total energy available as well as the maximum voltage and current.

Despite substantial prior work  on programmable matter and robotic origami~\cite{rus2018review}, current systems do not meet these requirements due to the well known scaling challenges of size, weight and power in the micro-robotics community~\cite{bergbreiter-review}. Specifically, designs requiring external magnetic fields or heating need close proximity to the source~\cite{ingestible-origami, miyashita2015untethered,magnetic-programmable-matter} and cannot operate in mid-air. Other heat actuated mechanisms like shape memory alloys~\cite{sma-origami-2010, robogami} and electrothermal polymers use large amounts of energy which would require substantially increasing size and mass to accommodate either a large solar array or a heavy battery. Additionally, these actuators have slow response times and can require over a minute to fold~\cite{printed-electrothermal}. Hygroscopic actuators that cause bending in response to humidity face similar drawbacks of long actuation times and are not electronically controllable~\cite{dandelion-morphing, humidity-glider}. Other designs have used motors combined with thread or gearing mechanisms to achieve folding but again require large and heavy batteries and actuation mechanisms~\cite{thread-origami, belke2017mori}. Piezo actuators are known to be highly efficient at small scale, however require high voltage boost converters to drive them which substantially increases mass and reduces efficiency~\cite{robofly,uwrobofly}. Similarly, origami designs driven by electrostatic actuators require even higher voltages of over 1 kV~\cite{electrostatic-origami}. These challenges are also highlighted by recently developed small folding robots weighing  approximately 10~g and cite difficulty scaling down in size due to their use of batteries which occupy  roughly half of the robot's surface area and consumes 40\% of its mass~\cite{tribot-nature}.

To solve these multi-faceted design challenges, we  analyzed  different actuation modalities to identify candidates that are compatible with solar power harvesting. Each solar cell produces a maximum of 2.8~V in bright sunlight. Although  this can be increased by connecting multiple cells in series,  achieving hundreds or thousands of volts would require a heavy and inefficient boost converter. Additionally, solar energy varies with factors like clouds and light intensity which requires a capacitor to buffer energy. Small capacitors also cannot tolerate high voltages due to dielectric breakdown, but do have low series resistance and can discharge current quickly. We therefore focused on actuators that require low voltages ($<$  10V) but higher currents. This suggests electromagnetic actuators or heat based actuation. We eliminate the latter due to their slow response times and high energy requirements. 

\begin{figure*}[t]
\vskip -0.3in
\centerline{ \includegraphics[width=1.\textwidth]{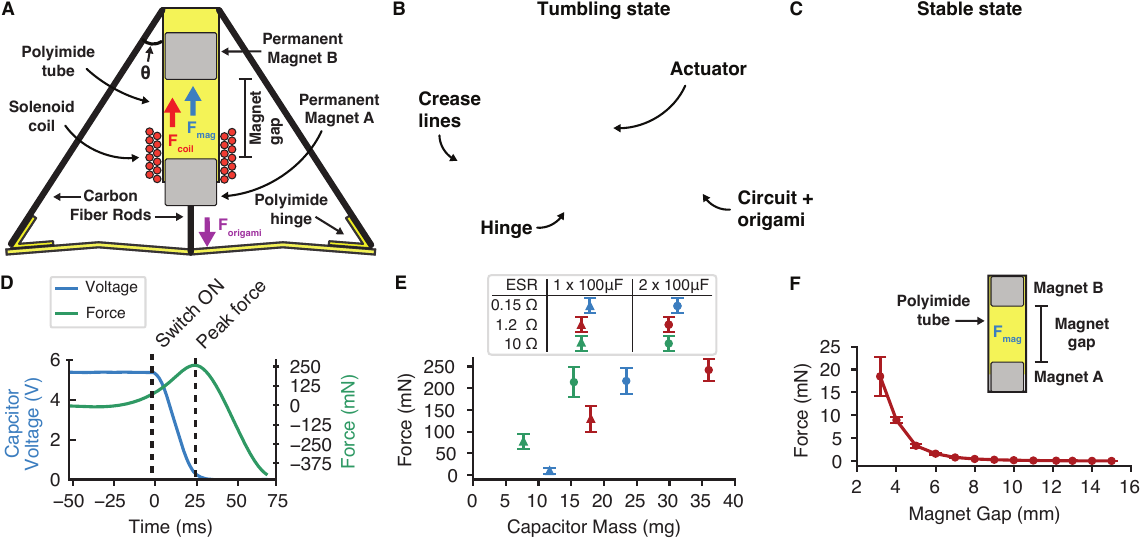}}
\vskip -0.15in
\caption{\textbf{Solar-powered actuator.} \textbf{A}) Diagram showing a cross sectional view with the components of our electromagnetic solenoid actuator. \textbf{B,C}) 3D models showing the actuator changing the shape of the origami from the flat (tumbling) state to the folded (stable) state. \textbf{D}) Waveform showing the voltage of the capacitor as it discharges into the actuator and the resulting force the actuator produces forces up to 250 mN when accelerating the magnet upward. \textbf{E}) Actuator force achieved for capacitors of different equivalent series resistance and capacitance versus component mass ($N=5$, $\pm\sigma$). \textbf{F}) Upward force exerted by the second magnet versus their separation gap ($N=4$, $\pm\sigma$).}
\label{fig:actuator}
\end{figure*} 

The most common electromagnetic actuators are continuously rotating motors, however we can leverage the properties of origami to simplify our actuation to a single pulse of linear motion. As long as the force on the major creaselines exceeds the energy barrier shown in Fig~\ref{fig:origami}B, it will snap into the other stable state. Thus, we designed a solenoid actuator that can provide the required linear motion. At a high level, the force of a solenoid depends on the
strength of the permanent magnet used and the current applied to the coil. The Lorentz force acting on the magnet can be expressed as~\cite{palak1, palak2}:
\begin{equation}
F_{Lorentz}\left(t\right)=B_{radial}\left(x\left(t\right)\right) I_{coil}\left(t\right) l_{coil} n_{turns}
\end{equation}
The term $B_{radial}$ is the magnetic field experienced by the coil which varies with the position of the magnet $x(t)$, $I_{coil}$ denotes the current applied to the coil,  $l_{coil}$ refers to the length of the coil and $n_{turns}$ is the number of windings. This lends itself well to our capacitor based energy storage scheme, as low series resistance capacitors can rapidly discharge high currents to create a motion pulse.

Fig~\ref{fig:actuator}A shows a diagram of the actuation mechanism which consists of a solenoid coil (30x3 turns array wound, 2.1 mm coil diameter) and a small 2.0~mm diameter neodymium magnet. We selected  the highest grade (N52) to maximize the strength of the permanent magnet. The magnet was constrained to moving up or down within a 2.1~mm diameter tube. The tube was made of an 12.5~{\textmu}m thick polyimide film to minimize the distance between the magnet and the coil. {Connecting a low series resistance capacitor to the coil results in a short, high amplitude pulse of current which accelerates the magnet upward causing the structure to fold and transition as shown in Fig~\ref{fig:actuator}B,C and Movie S6}. We note that these diagrams are shown with the tube oriented upward to better illustrate the components, however the microflier rotates in its tumbling state and falls with the tube oriented downward in its stable state.

We performed a series of benchmark experiments to characterize the force produced by the actuator itself prior to integration with the origami structure. As shown in Fig~S4, a magnet attached to a carbon fiber rod and flat reflector are placed in a polyimide tube with a solenoid coil at the base. The coil and tube are glued to a glass slide to keep them in place. A laser distance sensor (Keyence IA-030) shining down at the reflector and sampling at 1~kHz was used to measure the distance versus time as the magnet accelerated upward. The actuator was connected to the switch and capacitor circuit used on the origami microflier and probed with an oscilloscope (Tektronix MDO34) to measure the capacitor voltage shown in Fig~\ref{fig:actuator}D. The second derivative of the distance waveform and the mass of the magnet and carbon fiber rod were used to calculate the force. To achieve maximum force with minimal weight, we evaluated  a number of capacitors with different properties and plot the peak force they produce. The same measurements were repeated N=5 times for each of the capacitors shown in Fig~\ref{fig:actuator}E. We began by testing a single capacitor, and then proceeded  to test two capacitors of the same type in parallel. 

{The peak value of approximately 250~mN was more than six times greater than the force required to transition the structure.} The waveforms also show our actuator can be controlled electronically and responds within tens of milliseconds,  meeting the requirement for fast motion.  {Fig~\ref{fig:actuator}E further demonstrates the ability to produce sufficiently large pulses of force to transition with only tens of milligrams required for energy storage.}

Coupling this force to the structure however introduces additional challenges as we seek to replicate the bending required for state transition shown in Movie S4. The magnet and coil must be able to push against each other. The structure creases  must also be able to bend freely to transition, and as the structure folds inward, the perimeter of the structure shrinks. To address this, we  rigidly attached the magnet to the center of the origami structure with a carbon fiber rod, as the center point  experiences motion only in the Z direction. We also suspended the tube and coil above it using carbon fiber rods connected to the outer edges of the structure. The rods were rigidly glued to the top of the tube, but are attached to the origami structure with flexible hinges made of 12.5~{\textmu}m thick polyimide film as shown in Fig~\ref{fig:actuator}A,B.  {The rigid carbon fiber rods transfer force to the outer edges of the structure, but must also be able to bend at their attachment points to accommodate folding. The magnet produces a force in the Z direction, whereas the rods  push the structure inwards in the X and Y directions. At a high level, our design operates similar to an umbrella. Maximizing the force delivered to the structure requires optimizing the initial angle of the rods $\theta$ shown in Fig~\ref{fig:actuator}A (see Supplementary Methods  and Fig~S5)}.

In order to balance the large force required to transition the origami structure with the need for low power, we  added an additional magnet as a passive method for reducing the force required by the actuator. By placing an additional permanent magnet in the tube above the coil, as soon as the magnet attached to the origami accelerates upwards, it will experience an attractive force that will help pull it upwards. Additionally, once transitioned, the origami will remain in this state.

\begin{figure*}[t]
\vskip -0.3in
\centerline{ \includegraphics[width=1.\textwidth]{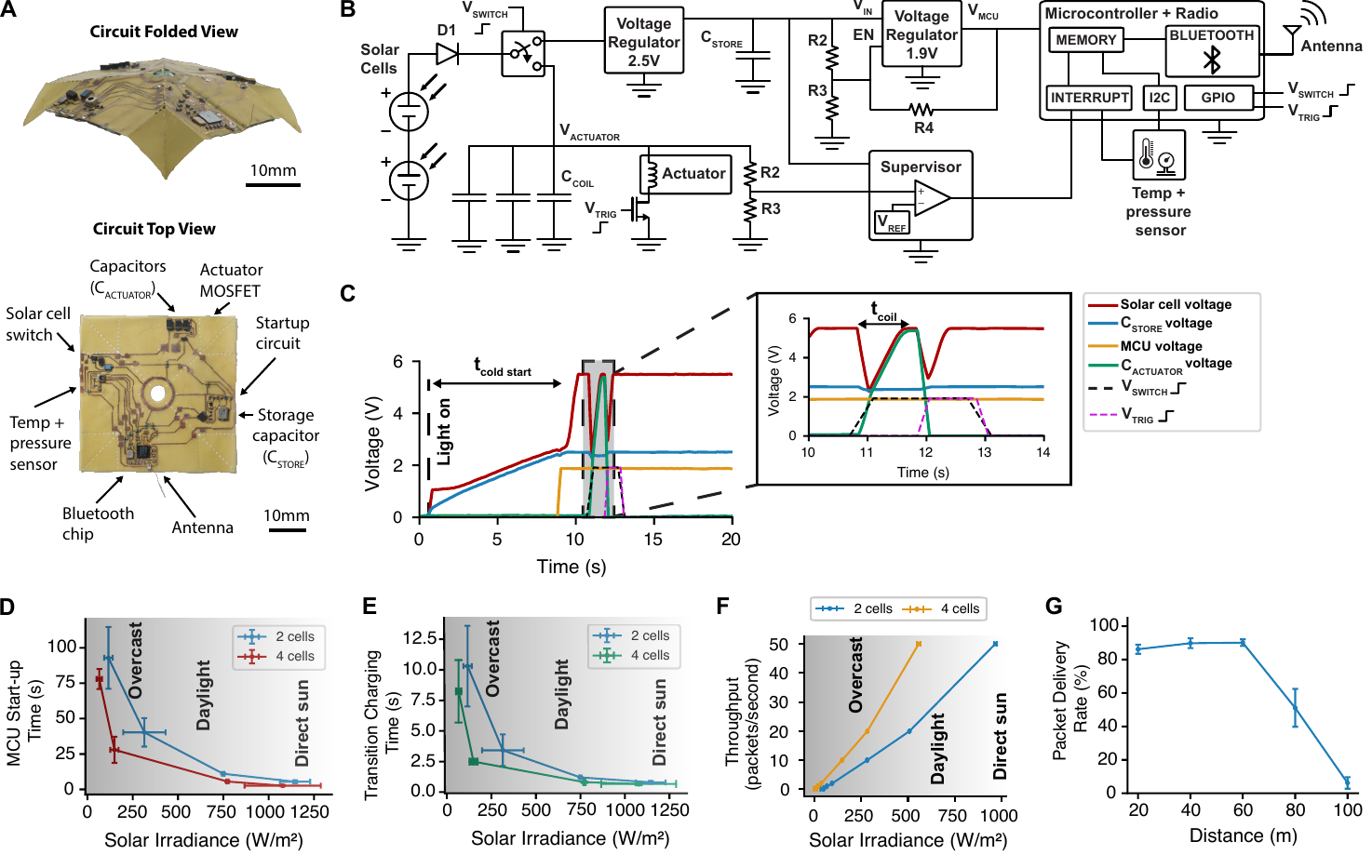}}
\vskip -0.15in
\caption{\textbf{Solar harvesting and wireless circuit.} \textbf{A}) Flexible circuit folded into 3D origami structure and labeled top view of major components divided into four regions for even weight distribution. \textbf{B}) Circuit diagram showing  power management for cold start, storage capacitors and different voltage domains, as well as the wireless microcontroller and sensor. \textbf{C}) Waveform illustrating microcontroller cold start followed by charging and triggering the actuator for transitioning the structure. \textbf{D}) Time required to cold start the microcontroller at varying light levels ($N \geq 3$, $\pm\sigma$). \textbf{E}) Time required to charge up the capacitors to transition at varying light levels ($N \geq 3$, $\pm\sigma$). \textbf{F}) Achievable Bluetooth throughput at varying light levels (N=3, $\pm\sigma$). \textbf{G}) Bluetooth packet delivery rate versus range (N=1000, $\pm\sigma$).}
\label{fig:circuit}
\end{figure*} 

We measured the attractive forces between our two 2.0~mm diameter N52 magnets empirically. To do this, we placed a polyimide tube on an approximately 3~cm tall plastic object to create a raised platform on a precision weight scale (Sartorius QUINTIX125D-1S) to make sure the magnets have no interaction with the scale and that metal components in the scale do not affect our measurement. We confirm magnetic effects to not affect the scale readings at this distance. We then placed our cylindrical magnet (Magnet A, 1~mm diameter, 2~mm height, grade N52) on the plastic surface and place the tube around it. We then glued another one of the same size magnets (Magnet B) to the end of a carbon fiber rod and lower it through the top of the tube using a micromanipulator. We recorded  the decrease in mass of Magnet A to determine the attractive force caused by lowering Magnet B, shown in Fig~\ref{fig:actuator}F. As expected, the attractive force increases non-linearly as the magnets approach each other. This allowed us to achieve a robust transition with less energy input to the coil. These results demonstrate a fully functional actuator compatible with solar-power harvesting that meets all of our design requirements.

\vskip 0.05in\noindent{\bf Solar harvesting and wireless circuit.} 
Programmably triggering our origami microflier to transition in mid-air requires an electronic circuit with multiple components for sensing, control, and power regulation. At the core of the circuit is a programmable microcontroller with an onboard Bluetooth radio (nRF52832, Nordic Semiconductor) which reads data from a temperature and pressure sensor (BMP384, Bosch) and can send a control command to the actuator to trigger it.   We fabricated the entire circuit directly on a flexible sheet of copper coated 12.5~{\textmu}m polyimide as shown in Fig~\ref{fig:circuit}A. This allows for integrating the electronics directly with the origami structure (see Methods). Additionally, this method enables scalable production of origami structures with integrated electronics using industry standard circuit fabrication techniques.

Running this device with  solar power however requires addressing multiple challenges. {To transition in mid-air, the solar cells must rapidly charge a lightweight energy storage element such as a capacitor to above 5~V for the actuator to generate a large enough force to transition our fully assembled origami microflier, as shown in Fig~\ref{fig:actuator}D}. However our microcontroller can only tolerate a maximum of 3.6~V which introduces the need for dual power regulation circuits to multiplex a single solar cell array and the ability to send control commands to these components (see Supplementary Methods  for embedded software details). Additionally, to operate for extended periods of time after deployment, these devices need to be able to cold-start without any stored energy due to the lack of an onboard battery. This is challenging as many microcontrollers have short, high current power spikes when turning on. This is often due to initialization procedures such as waiting for their clock oscillators to stabilize before they can run code to go into their low power modes.

To address these challenges, we designed the lightweight circuit shown in Figs~\ref{fig:circuit}A,B. Power was provided to the circuit by a lightweight solar cell array. To achieve the voltage required by our actuator, we used a minimum of two solar cells (5x5~mm, Microlink Devices) connected in series. For more robust operation in low light environments, we connected an additional two cell array in parallel. The output of the solar cell was connected  to a diode to prevent reverse current flow and then into a single pole dual throw switch. We used this switch to multiplex our solar cells between powering the microcontroller and charging the capacitors for the actuator. This strategy allowed us to rapidly charge up for transitioning using all available power and then use the  power for sensing and data transmission after deployment with a single lightweight component. The switch control signal used a pull-up resistor to keep it in the default state of charging the microcontroller for startup.

In order to sustain the microcontroller while charging the actuator, our microcontroller needed an energy storage capacitor C\textsubscript{store}. We selected a small 7.5~mF supercapacitor which can sustain the microcontroller while transmitting Bluetooth packets at a rate of 1 packet per second for up to 1 minute, even with no input power. However, this component could only tolerate a maximum of 2.6~V.  We placed a voltage regulator which acts as a limiter to prevent damage. We also used a modified version of the lightweight startup circuit presented in our prior work~\cite{dandelion-sensors} to enable robust cold-start. Briefly, the circuit used a voltage divider from the input to trigger the enable pin of a second 1.9V regulator when C\textsubscript{store} is fully charged. Upon startup, an additional high impedance feedback path kept the system on. We also note that to achieve robust startup we place an additional 100~uF tantalum capacitor in parallel with the larger 7.5~mF capacitor to help buffer the initial transient power spike. 

Fig~\ref{fig:circuit}C illustrates the full operation of the power harvesting circuit from cold-start to actuation. When first exposed to light with zero charge at t\textsubscript{0}, the microcontroller storage capacitor C\textsubscript{store} begins to charge. Upon reaching 2.5 V, the voltage regulator limits the value of C\textsubscript{store} whereas  the solar array increases to its maximum voltage over 5~V. Simultaneously, the startup circuit with the 1.9~V regulator detects that C\textsubscript{store} is fully charged and turns on to power the microcontroller. The time required to turn on the microcontoller,  t\textsubscript{cold start}, was determined by the light intensity. 

We evaluated the time required to charge each of the capacitors in our circuit outdoors. We placed  the fully assembled origami microflier on its side with its solar cells facing up towards the sun. We connected  wires between the solar cell outputs and power input to measure current with a multimeter (Fluke 289), and use 43~AWG wires connected to different points on the circuit to measure the waveforms shown in Fig~\ref{fig:circuit}C using an oscilloscope (Tektronix MDO34). We placed  a solar power meter (TES 132) next to the solar cells at the same angle to measure the incident solar power. We extracted the startup  and charging times from the oscilloscope waveforms to generate the plots in Fig~\ref{fig:circuit}D, which shows the microcontroller startup time across a variety of light conditions with 2 and 4 solar cells.

The microcontroller then begins running code and can sample its sensor readings, run an onboard timer, or wait for a radio signal to determine when to transition. Fig~\ref{fig:circuit}C shows the circuit operating with a fixed delay after which it sets the V\textsubscript{switch} signal high and begins charging C\textsubscript{coil}. The time required to fully charge C\textsubscript{coil} also varies with light intensity. We performed  measurements outdoors as explained above and plot the results in Fig~\ref{fig:circuit}E with 2 and 4 solar cells. This latency between charging and transition introduces a dependency  between the available sunlight and the minimum height from which the sensors can be dropped. The circuit then waits for C\textsubscript{coil} to charge which can either be implemented with a programmed delay, or using an interrupt from the supervisory circuit which detects when  C\textsubscript{coil} has reached maximum charge. Next, the circuit sets V\textsubscript{trig} to high which enables the actuator and transitions the structure. 

{Movie S7 shows the resulting end-to-end operation of the microflier transitioning completely untethered using solar power. Additionally, Movie S8 demonstrates the same end-to-end operation transitioning in mid-air when dropped outdoors from a ladder at a height of approximately 4~m.} The origami microflier begins falling in its tumbling state, and then transitions in mid-air to it's stable state and changes its descent behavior.

After deployment, the microcontroller  continues   sampling its sensors. As shown in Fig~\ref{fig:circuit}D, the circuit can cold start from zero charge even in low light conditions. To understand how often our device can transmit data, we performed additional outdoor measurements to determine throughput. Due to the complexity of reprogramming the miniaturized circuit during different sun conditions, we measured the power required for Bluetooth first and then measured the power provided by the solar cells. To measure  the achievable Bluetooth throughput versus light level we  programmed the microcontroller to transmit packets at different rates. We tested the maximum and minimum delays between packets allowed by the chip in its advertising mode. We powered the circuit with a source meter (Keithley 2470) and recorded the average current for a duration of two minutes. We note that shortly after startup, the circuit drew more power, but then settled to a steady state after approximately 1~min. Minimal change is seen after 2~min. 

To determine the light level required to achieve each transmission rate, we performed the same measurement described above using a multimeter connected between the solar cells and the circuit input during the startup phase.  We plot the results in Fig~S6 and use linear regression to generate a mapping between light level and currents. The plot shows a highly linear mapping for both two and four cells with $R^2 > 0.98$. We used this data to generate Fig~\ref{fig:circuit}F. The results show that in brighter conditions, the circuit harvests enough power to transmit  data at the maximum possible rate allowed by the Bluetooth radio chip, sending a packet every 20~ms. 

We also evaluated the distance at which we can decode Bluetooth transmissions in outdoor environments. We evaluated  Bluetooth range in an open field by placing the origami microflier on the ground in the grass, and placing a receiver at increasing distances. We utilized the microflier's onboard antenna, a chip antenna (Johanson 2450AT14A0100) with an 8~mm length of 41~AWG wire attached to the end to improve performance~\cite{airdropping,camera,livingiot} as the transmitter. For the receiver, we connected an nRF52832 development board to an 8~dBi patch antenna (L-Com, RE09P-SM) at a height of 2~m. For each trial we transmitted N=1000 packets with sequence numbers from the origami microflier and count the number of packets correctly decoded at the receiver to determine packet delivery rate in Fig~\ref{fig:circuit}G. Despite its small antenna, we observed a robust link with low packet error rates up to 60-70~m. This demonstrates the potential for a drone to  fly over and collect data from the devices at a high altitude using a Bluetooth receiver.

\vskip 0.05in\noindent{\bf Outdoor field evaluation.} 
In addition to  characterizing the components of our solar-powered origami microfliers we performed  outdoor field experiments to evaluate their real-world behavior.  {First, we  verified that the difference in falling behavior shown in Movies S1, S2, and S3 also occurs  outdoors.} To evaluate this, we constructed a mechanism to drop our microfliers from a drone (DJI Mavic Mini,  Fig~S7). The origami microflier is   place in a compartment below the drone and a remotely triggered servo opens the compartment. {As shown in Movie S9, our origami microflier is dropped from the drone at a height of 20~m.} The video shows it begin falling in its tumbling state and then transition in midair to change its falling behavior to stable descent. {At the beginning of its fall the flier passes through a region affected by drone induced airflow~\cite{drone-flow1,drone-flow2,drone-flow3}. We observed in Movie S9 that this appears to have minimal effect with the flier falling straight downward, potentially with some downward acceleration due to downwash. We note that other drone configurations that induce more turbulent flow in this region could even help scatter the fliers laterally.}

\begin{figure}
\vskip 0.05in
\centerline{ \includegraphics[width=0.48\textwidth]{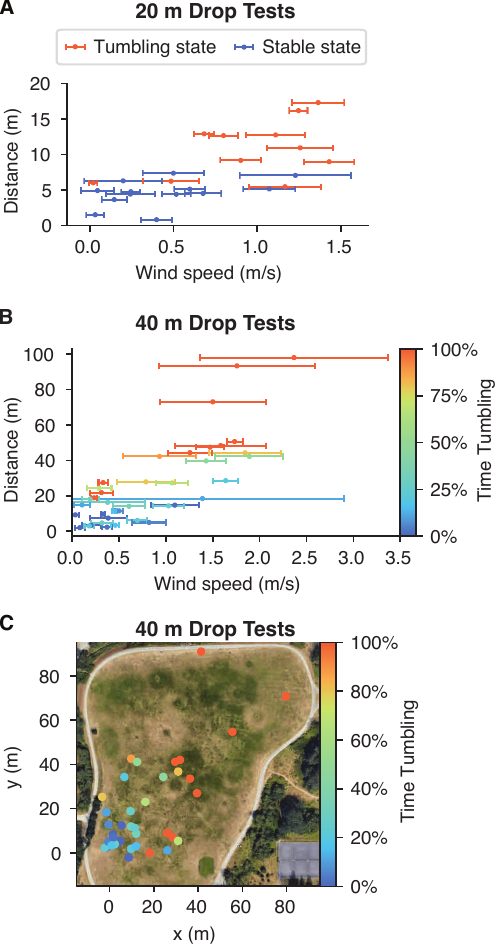}}
\vskip -0.15in
\caption{\textbf{Outdoor drop tests.} \textbf{A}) Outdoor drop tests from 20 m showing a comparison of distance traveled in the tumbling state and the stable state across wind conditions. $N=24$ trials (error bars indicate $\pm\sigma$, $N\geq 10$ wind measurements). \textbf{B}) Outdoor drop tests from 40 m showing distance traveled when the microflier transitions in mid-air ($N=38$ drop trials, error bars indicate $\pm\sigma$, $N\geq 30$ wind measurements). The colorbar indicates the time spent in the tumbling state before transitioning to the stable state. \textbf{C}) Data from B) shown as a map of landing locations, drone liftoff from (0, 0)}.
\label{fig:outdoor1}
\end{figure}

{We repeated these experiments in a range of altitudes and wind conditions as shown in Movie S10 to evaluate dispersal distance. We used a hot wire anemometer sampling at a rate of 1~Hz (405i, Testo) to measure the lateral wind speed during these trials. We placed the anemometer at a fixed height of 2~m oriented in the direction that the microflier traveled due to the challenges of measuring ambient wind speed at varying altitudes and trajectories in the field. We also measured the distance from the drop location to the microflier's landing site. We combined  these data to investigate the effect of lateral wind and dispersal distance seen in controlled experiments in Movie S3.}

\begin{figure*}[t]
\vskip -0.3in
\centerline{ \includegraphics[width=.8\textwidth]{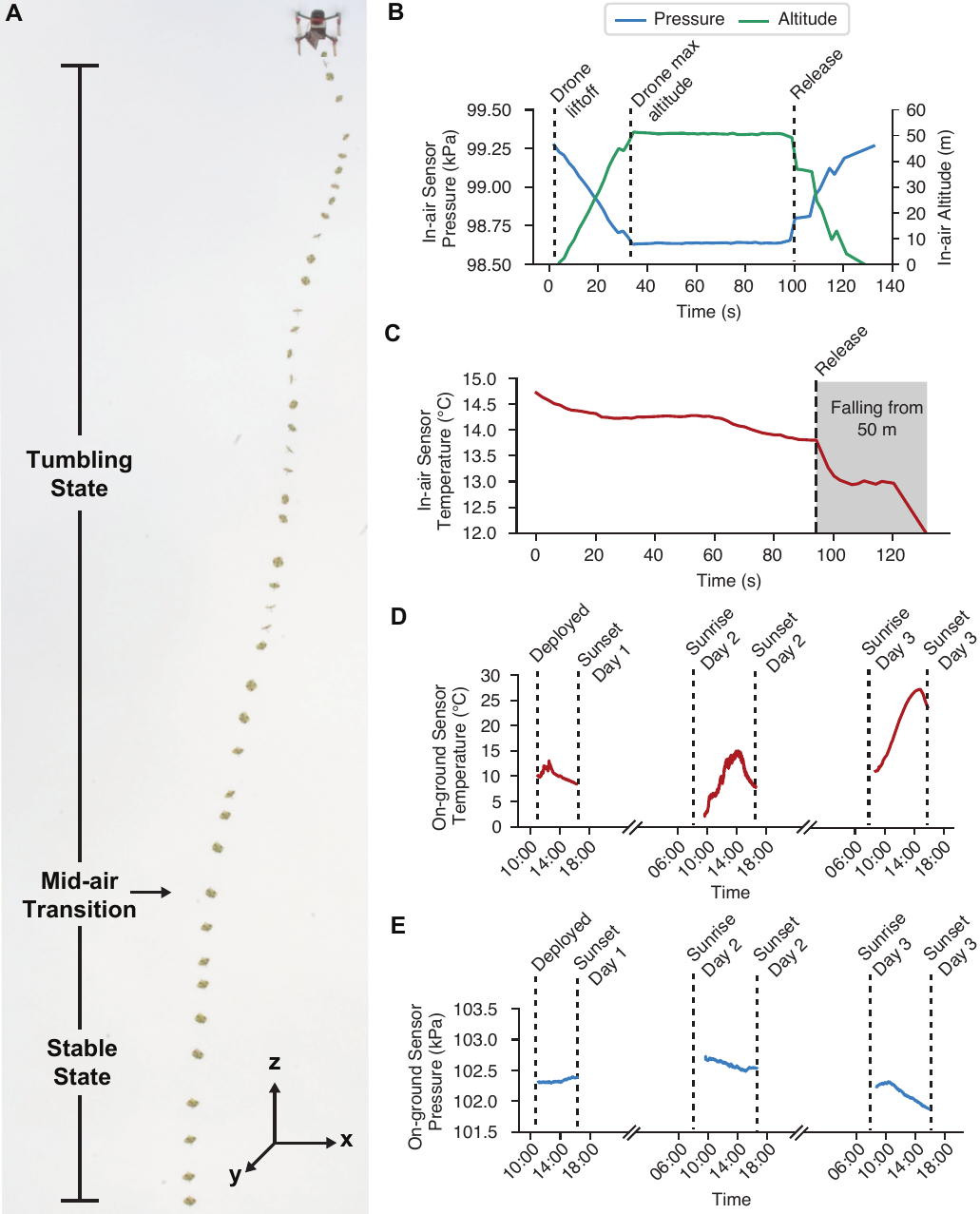}}
\vskip -0.15in
\caption{\textbf{In-air and on-ground microflier  measurements.} \textbf{A}) Images showing our origami microflier falling from a drone and transitioning from its tumbling to stable state in mid-air. \textbf{B}) Real-time pressure sensor readings sent via Bluetooth showing the altitude of our origami when dropped from 50 m. \textbf{C}) Real-time temperature sensor readings sent via Bluetooth showing the temperature as our sensor falls. \textbf{D}) Temperature readings from a 3 day outdoor deployment demonstrating that the microflier  cold-starts daily with harvested power. \textbf{E}) Pressure readings from the 3 day outdoor deployment.}
\label{fig:outdoor2}
\end{figure*}

Fig~\ref{fig:outdoor1}A shows a comparison of the distances traveled in tumbling and stable states versus wind speed when dropped from a height of 20~m in an open field. The data shows a  division between the two states, especially at higher wind speeds, confirming that the observations in Movie S3 hold outdoors in higher altitudes as well. This is intuitively because  the rotation of the sensor in the tumbling state helps maximize the area exposed to lateral wind gusts.

We also performed a series of experiments  transitioning the origami microfliers in mid-air. To change the shape at different altitudes, we programmed our microfliers to transition a fixed time delay after receiving a trigger command sent over Bluetooth. The origami microfliers were also programmed to  continuously broadcast  Bluetooth packets with values of their onboard counter as well as temperature and pressure readings. In our implementation, we could use the pressure sensor, a timer, or a Bluetooth command to trigger the transition. We used the same drone and wind measurement setup described above and performed drop experiments in an open park space from an altitude of 40~m. Trials were performed sequentially to facilitate recording data.

Fig~\ref{fig:outdoor1}B shows the distance traveled versus wind speed and Fig~\ref{fig:outdoor1}C shows the same data represented as a map of microflier landing locations. Additionally, the colorbar indicates the amount of time spent in the tumbling state versus the stable state. For example, 100\% would indicate the sensor was constantly tumbling and was not programmed to transition whereas 0\% would indicate it was programmed to immediately transition and fell only in the stable state. Values in between 0\% and 100\% indicate a mid-air transition. We observed that similar to Fig~\ref{fig:outdoor1}A, the longer the sensor spent in the tumbling state, the farther it traveled. {Further, across microfliers that spent a similar amount of time in their tumbling state, those that experienced higher wind speeds traveled longer distances. We note that although the wind varied across our individual experiments, when dropping multiple devices programmed to transition at different times or altitudes they will experience the same conditions and travel different relative distances.} These plots  demonstrate that the transitioning mechanism works in real world environments and can achieve our goal of varied dispersal distances.

Due to their low mass and terminal velocity, our origami microfliers were not damaged by physical impact with the ground. Our design however requires that they land with their solar cells facing upward. To evaluate this, we performed a series of experiments dropping our origami microfliers onto a grass surface from a height of 2~m. We observed that in the tumbling state, the microfliers land upright only 52\% of the time (N=50). However, in their stable state they land upright 87\% of the time (N=50). In the stable state we observed that an even greater number (96\%) initially landed upright but either bounced or collided with an object causing them to flip. This shows  that by transitioning to the stable
state before landing we can increase the probability of our devices landing  upright to harvest power.

Fig~\ref{fig:outdoor2}A illustrates  the trajectory of an origami microflier transitioning in mid-air when dropped from 15~m. Although this 2D image makes it difficult to visualize the difference in lateral distance traveled along the Y-axis, the trajectory  shows the device tumbling until it transitions and stabilizes in mid-air. Figs~\ref{fig:outdoor2}B,C also show data recorded from Bluetooth transmissions in real time during a drop from 50~m. Fig~\ref{fig:outdoor2}B shows the in-air sensor data as  raw pressure measurements  converted to altitude. The plot shows  the altitude increasing as the drone lifts off and ascends to its target height of 50~m. Upon release, the microflier begins to fall to the ground and shows a corresponding decrease in altitude. Fig~\ref{fig:outdoor2}C shows the temperature data from the microflier as it falls from the drone, again demonstrating the ability for the solar array to power our microfliers in mid-air and the potential for taking atmospheric measurements at different altitudes.

We also evaluated performance of the microflier on the ground post-deployment to verify the circuits can cold-start and operate on solar power for extended periods of time. 
The data was collected by a computer receiving the Bluetooth measurements.  {Fig~\ref{fig:outdoor2}D,E show the results over the course of three days, during which the sensor turns off at sunset and successfully cold-starts shortly after sunrise.} During this time it was able to send 24,000 Bluetooth packets and operate continuously for 6~hours per day. We note that the operational time was limited by the number of sunlight hours at the deployment location in Seattle, WA in December. We performed additional measurements to evaluate sensor performance in comparison to a reference device (see Supplementary Methods  and Fig~S8).
\section*{Discussion}
Here, we designed  solar-powered origami microfliers that can change their shape in mid-air to vary their dispersal distance. Our key observation is that leaf-out origami structures exhibit distinct falling modes in their two  states: a tumbling behavior conducive to wind dispersal and a stable descent state less affected by wind. We  co-designed a light-weight, low-power actuator and solar power harvesting circuit that enables the  microflier to change its shape  in mid-air using  solar power harvested outdoors.  Our design uses a programmable microcontroller that makes it extensible to adding other sensors for a wider range of environmental monitoring applications. 

Adding more payload, however,  requires additional consideration to maintain a robust difference in falling behavior between the two origami states. We find that the device is sensitive to weight distribution and that adding too much weight on one unit cell can cause asymmetry and thus, flipping in both  states (see Supplementary Methods). This can be addressed by adding  balancing masses to ensure that weight is uniformly distributed across all cells.  Further, the corners of our origami structures are not reinforced with carbon fiber and are flexible which could also affect stability with larger payload.

Our origami microfliers  support bi-directional radio connectivity via Bluetooth. This allows them to not only communicate with a  base station, but potentially communicate amongst each other and form peer-to-peer or mesh networks. This presents multiple opportunities to both increase operational range through multi-hop communication, while also presenting routing and scheduling challenges due to variability in solar power  on  terrains with complex geographic features~\cite{intermittent}.  We also observe that in brighter conditions, our actuator charges fast enough to be triggered repeatedly. Modifying the design for bi-directional transitions between the states could enable more precise control over falling behavior and even motion such as jumps after falling to the ground. Further, our actuator, circuit, and fabrication methodology to pattern electronics directly onto the folding structure can be applied broadly to the field of origami robots enabling a range of miniaturized battery-free  designs.

Finally, although our microfliers can enable   field deployments of environmental sensors, we must also consider their environmental effects at their end of life and their potential to create waste. One solution is to recollect the devices after deployment. {The magnets in our actuators present a means for automated collection by sweeping a magnet over the deployment area. We evaluate this using a magnetic sweeping device used to collect nails and other metallic objects on construction sites as well as a small neodymium bar magnet.  As shown in Fig~S9A-C, even these relatively small magnets can collect our microfliers from distances of 5~cm, and adding an extra onboard magnet extends this to 6~cm. These experiments demonstrate the potential for automated collection for example by pulling an electromagnet behind a tractor in agricultural deployments.} 
Our onboard radio could also be used for localization and presents opportunities to use  recent advances in the  sensor networking communities~\cite{localization,bleloc}. {An alternative method that is more attractive for remote and difficult terrain would be to incorporate biodegradable materials in the design of our microfliers \cite{mouse}}. Our design already eliminates batteries, and we can build on this to use biodegradeable materials such as cellulose for the structural components~\cite{humidity-glider}. We hope that future advances in sustainable materials paired with innovation in micro-robotic systems could enable this vision.
\section*{Materials and Methods}

\noindent{\bf Origami circuit fabrication.} We fabricate our origami structures by folding sheets of polyimide films (Dupont Kapton and DuPont Pyralux AC121200E) with the pattern in Fig~\ref{fig:origami}A. We choose the number of unit cells and use a custom python script to generate a tessellation of unit cells with the appropriate angle $\alpha$. Next, we choose a film thickness and cut pattern. We tune the structure's initial folding angle by using a small number of holes along the boundary creases and adding more cuts along the main and sub creases. The stiffer boundary crease dictates the initial folding angle and energy barrier between states. Combined, these factors determine the energy required for transition.

We use laser micromachining (LPKF U4) to cut out the shape of the origami and make the crease cuts. {We choose four unit cells for our microfliers using data in Fig~\ref{fig:origami}D,E and Fig~S10 which indicate this design has the highest lateral displacement which is correlated to the greatest number of rotations during its descent, and the lowest terminal velocity. Additionally this design is symmetric which simplifies weight distribution and fabrication}. We choose a side length of 39~mm based on experiments showing smaller prototypes did not exhibit different descent behaviors. This size achieves our sub-gram target mass and provides sufficient area for the electronics.

We further demonstrate direct patterning of functional circuits onto the foldable origami to create our final microfliers shown in Fig~\ref{fig:fig2}. We use a copper coated film (12~{\textmu}m copper, 12~{\textmu}m polyimide, DuPont Pyralux \\
AC121200E). We first cover the copper surface in an ink mask. Next, we use the laser to raster away the mask in regions around the desired pads and traces leaving the copper exposed. We then make the crease and boundary cuts. We etch the exposed copper using Ferric chloride and remove the remaining ink using acetone or isopropanol to create the final circuit. Components are then manually placed under a microscope and soldered using a hot plate at 285~°C. This creates a fully functional circuit on a flat, flexible sheet. This process is similar to commercial flexible circuitboard fabrication which can be used to scale up production.

To maintain rigid origami, we attach a carbon fiber root structure as shown in Fig~\ref{fig:fig2}C to reinforce the faces. We first create 110~{\textmu}m thick carbon fiber layups ($0$°,$45$°,$-45$°,$0$°, Toray M46J) by laminating the layers together in a heat press (80 psi at 150°C for 90 minutes). We use the same laser micromachining procedure to cut out eight sections in the desired patterns matching the origami faces. We choose the root pattern to add rigidity with minimal mass. We attach the subsections of the root structure together by placing them on a piece of 50 {\textmu}m thick PET tape (Gizmo Dorks) to create a flexible hinge. The PET tape is patterned with cuts allowing it to bend more easily and act like a hinge at the fold of the structure. This allows for the root structures to fold at angle $\psi$. We attach the root structure to the leaf-out circuit in its flat state using cyanoacrylate (CA) glue. We then  apply the origami folds shown in Fig~\ref{fig:origami}A to create the folded origami circuit as shown in Fig~\ref{fig:circuit}A. The solar cells are placed on pieces of kapton tape and attached to the opposite side of the structure. The solar cells (2-4x 5x5~mm cells, Microlink Devices) are manually wired together and soldered to the remainder of the circuit using 43~AWG wire.

\noindent{\bf Actuator fabrication.}
After creating the origami structure and circuit, we fabricate our miniaturized electromagnetic actuator and attach it to the structure. We roll a sheet of polyimide film (12.5~{\textmu}m, Dupont Kapton) into a tube (2.1~mm diameter, 10.5~mm length) to restrict the motion of the magnet. This tube is designed to match the inner diameter of our solenoid coils (2.1mm diameter, 2.1mm height, 30x3 turns array wound, Golden Eagle Coil \& Plastic Ltd). We insert the tube into the coil and use CA glue to attach it at one end of the tube. We then attach a small neodymium magnet (1.0mm diameter, 2.0mm height, grade N52) to a carbon fiber rod (0.25mm diameter, 8.0mm length) and glue it to the center of the root structure to avoid  interference with the creases of the hinge. We then glue a  second magnet-rod component inside the tube such that it is suspended approximately 8.0~mm above the other magnet when the structure is in its flat, tumbling state. These measurements are empirically determined from Fig~\ref{fig:actuator}F, and can be adjusted depending on the strength of the attractive force between the two magnets and the required transition force. 

\noindent{\bf Statistical tests.} Relevant uncertainties and statistical methods are noted in the corresponding figure legends.

\vskip 0.1in\noindent{\bf Acknowledgements.} The authors thank Hiromi Yasuda, Jinkyu Yang and members of the Laboratory for Engineered Materials and Structures for advice and helpful discussions on origami design. We also thank Nicole Sullivan for assistance with outdoor drop test experiments. 
 
\vskip 0.05in\noindent{\bf Funding.} The researchers are funded by a Moore foundation fellowship, National Science Foundation (ECCS-2054850), National GEM Consortium, Generational Google Scholar Fellowship Program, Cadence Fellowship Program, Washington NASA Space Grant Fellowship Program, and SPEEA ACE Fellowship Program. 
 
\vskip 0.05in\noindent{\bf Data and material availability.} All data needed to evaluate the conclusions of the paper are available in the paper or the Supplementary Materials. The data for this study has been deposited in the following database: 10.5281/zenodo.8247362.

\vskip 0.05in\noindent{\bf Author contributions.} 
K.J. and V.A. designed and fabricated the origami structures; K.J., R.V., V.A. and V.I. fabricated circuits and actuator parts. T.E. and V.I. developed the embedded software. K.J., V.A., V.I., A.F., A.A., S.G. and S.F. designed the experiments. K.J., V.A., R.V., D.Y., A.F., A.A. and V.I. conducted experiments and characterized system performance; K.J., V.A., A.F., V.I., D.Y. analyzed data and generated figures; V.I., K.J., A.A. and S.G. wrote the manuscript; S.G., V.I., A.A. and S.F. edited the manuscript. 

\vskip 0.05in\noindent{\bf Competing interests.} S.G. is co-founder of Jeeva Wireless and Wavely Diagnostics. The other authors declare that they have no competing interests.

\bibliographystyle{ACM-Reference-Format}
\bibliography{scibib1}

\clearpage
\section*{Supplementary Information}

\subsection*{PIV analysis}

\begin{figure*}[t]
\vskip -0.3in
\centerline{ \includegraphics[width=0.99\textwidth]{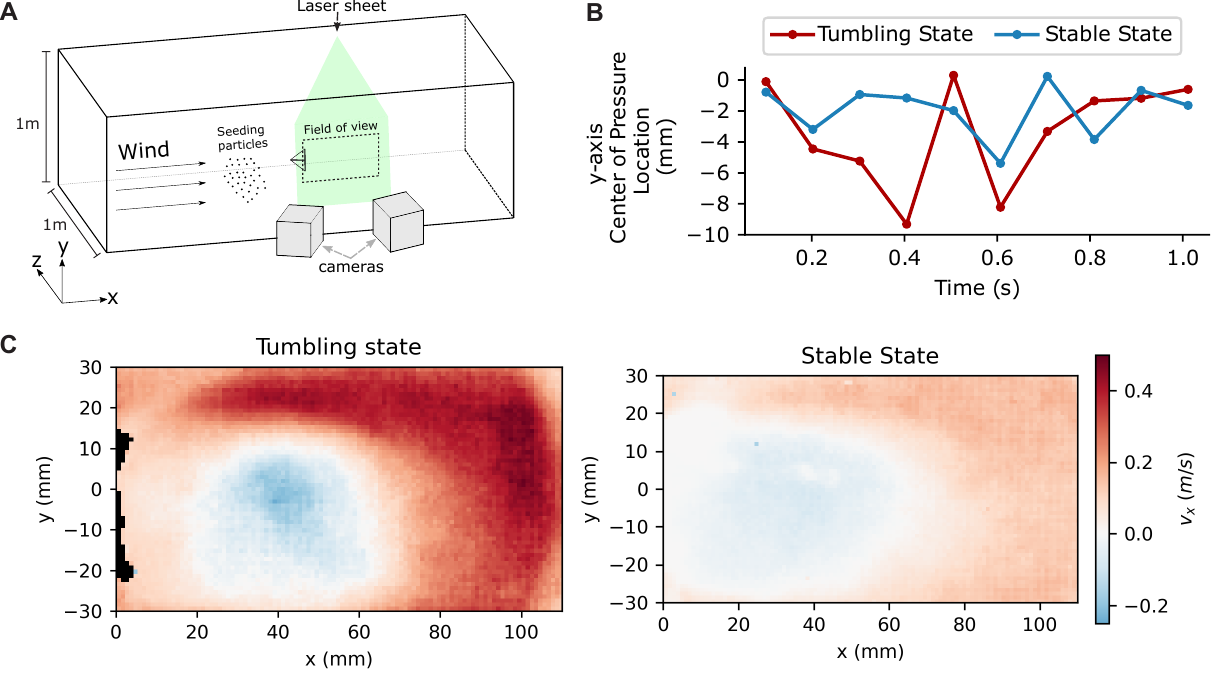}}
\vskip -0.15in
\caption{\textbf{Fluid Analysis.} \textbf{A}) Diagram showing the PIV measurement setup in a square cross-section wind tunnel (1 m $\times$ 1 m). A fog machine (ADJ VF400) placed 120 cm upstream produced the seeding. Stereoscopic images were collected at a frame rate of 890 Hz  with a field of view measuring 13 cm (in the streamwise direction) by 8 cm (in the spanwise dimension). Illumination came from a solid-state dual-head laser (TerraPIV, Continuum, San Jose, CA, USA) emitting short pulses of visible light at 527 nm synchronized with the cameras (Phantom v641, Vision Research, Wayne, NJ, USA). Data was analyzed using PIV image analysis software (Insight 4G, Version 11.1.0.5, TSI Incorporate). \textbf{B}) Movement of center of pressure along the y-axis versus time. In the tumbling state the center of pressure shows substantial and rapid oscillation from side to side causing the flier to tilt and begin tumbling. \textbf{C}) Time averaged 2D velocity profiles visualizing the magnitude of the flow field in the wake immediately above the flier in its stable and tumbling states. The origami face is placed at x=0 with small amounts protruding into the plot in the Tumbling state figure. We note that the averaged velocity increases beyond x=100 mm but is not visualized in the figure due to the limited field of view chosen to investigate the flipping dynamics of the flier.}
\label{fig:piv}
\end{figure*} 

Particle Image Velocimetry (PIV) measurements were taken in an open loop wind tunnel~\cite{wind-tunnel1} as shown in Fig~\ref{fig:piv}A. The wind tunnel has a square cross-section measuring 1 m x 1 m. The free stream velocity is produced by an array of electric fans controlled by constant DC voltage. The robotic flier was held with its plane perpendicular to the free stream flow at the end of a thin ($2\times 2$ mm) aluminum beam. A propylene glycol fog machine (ADJ VF400) was used for seeding the flow and placed 120 cm upstream of the robotic flier.
 
{Stereoscopic particle image velocimetry (PIV) measurements were collected at a frame rate of 890 Hz. Illumination came from a solid-state dual-head laser (TerraPIV, Continuum, San Jose, CA, USA) emitting short, $O(10^{-8})$s, pulses of visible light at 527 nm. The laser was synchronized with two high-resolution, high-speed cameras (Phantom v641, Vision Research, Wayne, NJ, USA). The flier and laser plane were arranged so that the laser plane was oriented in the streamwise and vertical direction (x-y) as shown in Fig~\ref{fig:piv}A to visualise the flow in the wake of the flier. Light from the laser plane, scattered by the seed particles in the flow, was captured by the two cameras with a 13 cm (in the streamwise direction) by 8 cm (in the spanwise dimension) field of view. The laser plane was located at the plane of symmetry (center) of the model, and a total of 900 image pairs were recorded. The time between two images used to compute the velocity field is $10^{-3}$s. The data was processed using PIV image analysis software (Insight 4G, Version 11.1.0.5, TSI Incorporated) using Gaussian correlation fitting, multipeak detection, and gradient-based iterative correlation for subpixel resolution. The velocity field in the wake of the flier was calculated with a recursive grid engine with a first pass on 64x64 pixel windows and a second pass on $32\times 32$ pixel windows.}

{The origami flier operates as an object falling at a constant speed, normal to its main plane. At 16 cm$^2$ square cross section and falling at approximately 1 m/s  terminal velocity (depending on configuration). The aerodynamics are those of a blunt body at Reynolds number 400-2600. Its wake is characterized via PIV measurements using the method described above. We present the results of these measurements in Fig~\ref{fig:piv}B,C.}

{The aerodynamic torque was characterized in Fig~\ref{fig:piv}B by taking the moment of the velocity profile, averaged over periods shorter than the instability of the wake (vortex shedding), with respect to the center line of the flier. The magnitude of the aerodynamic torque is represented by the location of the center of pressure with respect to the center of the flier. The data shows it is substantially higher in the flat, tumbling state than in the stable state, and its switching frequency is also higher. This represents the main mechanism which causes the flier to start tilting to one side and begin tumbling. After rotating 90$^\circ$ from a stable fall in which the flier experiences high drag due to its plane being normal to gravity (and the fall velocity), its drag decreases as its plane becomes aligned with gravity and the fall velocity.}

{In its flat, tumbling state the flier presents sharp edges to the incoming air flow, when the flier is falling with its plane perpendicular to the flow (gravity). Fig~\ref{fig:piv}C compares the wake of the two states by visualizing the mean flow over a 1~s interval as color contours of constant velocity deficit. The plot shows a wide wake and the extent of the wake momentum deficit resulting in widely different drag coefficients. In comparison flier presents beveled corners to the flow in its folded stable state, resulting in narrower wake, with a vortex forming closer to the center of the flier, resulting in a more symmetric wake and lower aerodynamic torque with respect to its center of mass. Thus, after transition to its folded state the filer is more stable: less prone to turning its plane from falling perpendicular to gravity to aligned with gravity.}

{In a Lagrangian sense, the flier falls under the effect of gravity and moves horizontally under the effect of lateral wind. It falls with a high drag coefficient and therefore low terminal velocity, as long as the flier’s plane is predominantly normal to the direction of relative wind (a combination of the wind direction and the fall velocity). Under the effect of aerodynamic torque, the tumbling flier will turn and align its plane with the direction of the relative wind, resulting in a much lower drag coefficient and much higher fall velocity. Additionally, while aligned with the relative wind direction, there is no aerodynamic torque to keep turning it away from alignment. While aligned with the relative wind direction, there is no aerodynamic torque to keep turning it away from alignment, however its angular momentum and fluctuations in wind will perturb this motion, returning the flier to an orientation with its face normal to gravity. In contrast, fliers in the folded, stable state will tend to fall with their plane approximately perpendicular to gravity with a low drag coefficient in the lateral direction throughout their descent.}

{Statistically, the direction and speed of the wind will affect the Lagrangian trajectories of the fliers randomly; however the dynamics of the two states will cause a relative difference in behaviors. The tumbling flier will turn and alternate aligning its plane normal to the ground and the direction of lateral wind causing a changing drag coefficient and higher horizontal dispersal while fliers in the stable state will tend to fall with higher vertical and lower horizontal drag. Thus, transitioning fliers between states allows for control of their trajectories and a means to separate a population of fliers. Changing the time to transition can allow multiple sampling populations with distinct residence times in different parts of the atmosphere and varying lateral dispersal as shown in Fig~6.}

\subsection*{Leaf-out origami kinematic simulation}
Using the kinematic models developed in~\cite{leafout1,leafout2}, we analyze the energy required to fold the structure to an angle $\psi$. We model each crease line $j$ as a torsional spring with spring constant $\kappa_j$ and folding angle $\rho_j$, defined as the complement of the angle between adjacent surfaces connected by that crease. Positive angles represent valley folds and negative angles represent mountain folds. We compute the potential energy $E$ required to fold the structure to angle $\psi$ with respect to the lowest energy state of the system, where $\bar{\rho}_j$ is the rest folding angle of $j$th crease line. We sum the individual energy of each of the folds up to $N_{Total}$, which represents the total number of crease lines in the leaf-out origami:

\begin{linenomath}
$$
	E =\frac{1}{2}\sum\limits_{j=1}^{{{N}_{Total}}}{{{\kappa }_{j}}{{\left( \rho_j -{{{\bar{\rho }}}_{j}} \right)}^{2}}},
$$
\end{linenomath}
{Because our structures are made of a uniform sheet of material, we assume an identical torsional stiffness for all creaselines ($\kappa_j = \kappa_\theta \text{ for } j=1,\dots, N_{Total}$). We use the relationships developed in prior rigid origami models~\cite{rig-orig2,leafout1,leafout2} to numerically solve for the individual fold angles $\rho_j$ as a function of $\psi$. We implement this using a custom Python simulation visualization framework developed in the authors prior work~\cite{leafout1}.}

\subsection*{Actuator attachment design considerations}
The solenoid coil is used to produce a force that pulls the magnet into the tube. In order to couple this to the structure however we need a way of attaching these points to the origami structure without restricting its ability to fold. Potential attachment points include the center and outside edges. Our magnet accelerates along the Z axis through the tube similar to an umbrella, and we suspend the tube from the outside edges of the structure using a set of four carbon fiber rods. These can be attached rigidly at one end, but require the ability to flex with the structure at the other. To achieve this we create hinges using small pieces of polyimide film (12.5~{\textmu}m, 1x3~mm) and glue them to one end of the rods. We select rods with a length of 30~mm and thickness of 0.25~mm. Each of these four rod-hinge components are glued to the four corners of the origami structure near the sub crease lines without circuit components attached as shown in Fig~4B. We then fold the hinges such that the rods attach to the tube at an angle $\theta\approx 45^{\circ}$.

\begin{figure}
\vskip 0.05in
\centering\includegraphics[width=1.0\linewidth]{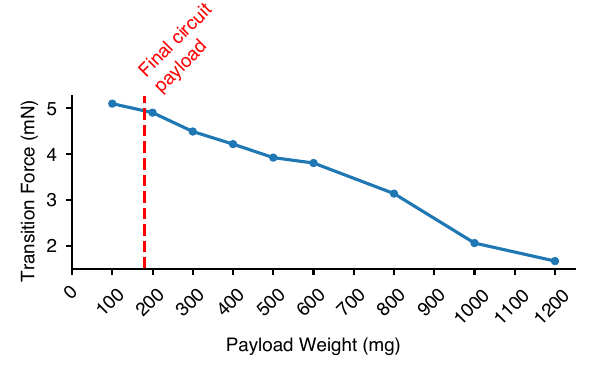}
\vskip -0.15in
\caption{\textbf{Origami transition force versus payload.} Force required to transition an origami structure (12.5~{\textmu}m Kapton, 15\% holes cut) with added payload but without the root structure  to simulate circuit components. A red dotted line indicates the weight of our circuit design which has minimal effect. The data shows a decrease in force required to transition as payload mass approaches and exceeds 1~g, highlighting the need for minimizing circuit mass to maintain a high enough force to present false transitions in mid-air. Adding the root structure further increases the required transition force.}
\label{fig:payload}
\end{figure}

We note that the length of the tube, rods, and angle play an important role in determining how much of the actuator's force along the Z axis is transmitted to the structure through the angled rods. As shown in Movie S4, we seek to bend the structure by forcing the center point upward and pushing the outer edges downward. The tube is rigidly coupled to the magnet and coil, and the rods are connected to the edge of the structure at an angle $\theta$. Fig~\ref{fig:force-diagram} shows a diagram with the desired Z component of the force we seek to maximize $F_{rod_z} = F_{rod}\,  cos(\theta$). Decreasing the angle $\theta$ by increasing the tube length results in an increased net-force at the tips of the rods. Given $\theta_1 < \theta_2$, $F_{rod}\,  cos(\theta_2) < F_{rod}\,  cos(\theta_1)$. Increasing the tube length however increases total size and mass. This also affects the structure as increasing payload decreases the force needed to transition as shown in Fig~\ref{fig:payload}.

{\subsection*{Assembly process and automation potential}
We illustrate the assembly of our fliers pictorially in Fig~\ref{fig:flow-diagram}. Our current prototypes are manually assembled and use low batch prototyping methods to enable rapid design iteration. Below we discuss potential techniques for scaling up production. The main body of our origami structure is created using standard flexible circuit materials and processes. This enables a substantial fraction of our device to be produced using existing industrial processes that are performed at scale for fabrication and assembly of flexible circuits. Printed circuit assemblies commonly use lamination to attach layers together. Adding an additional lamination step would enable the carbon fiber and PET to be attached during this process as well. The assembly could then be folded by pressing in a mold or using other techniques for industrially fold paper. We note that while our design uses handmade polyimide tubes in our initial prototypes, these are also commercially available as are solenoid coils. Additionally, while we use cylindrical carbon fiber rods in our prototypes, the whole rod and flexible linkage assembly could be produced as a single assembly using smart composite microstructures~\cite{robofly-fab, camera} which we have used in prior work and are common in the manufacturing of insect-scale robots. This leaves gluing the 3D assembly together as the final step. This could be automated by depositing drops of glue at the attachment points and lowering the tube and rod structures onto the circuit.}

{\noindent\textbf{Cost estimate.} We estimate the component cost for origami microfliers produced at scale to be approximately \$10.94 (excluding the solar cells), as shown in Fig~\ref{fig:cost}. We estimate the cost of the microflier without including the solar cell components because the number and quality of the solar cells can be adjusted based on the application requirement.}

\subsection*{Weight distribution}

\begin{figure}
\vskip 0.05in
\centering\includegraphics[width=.75\linewidth]{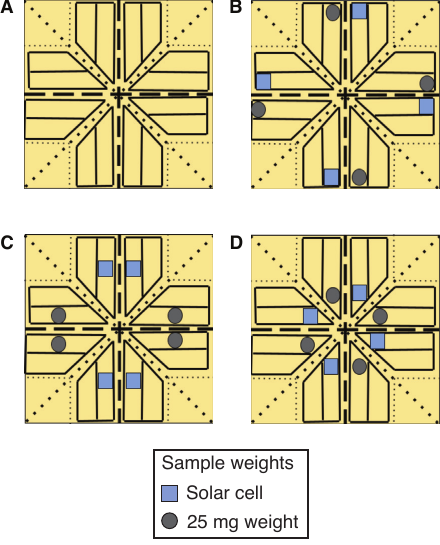}
\vskip -0.15in
\caption{\textbf{Weight distribution tests.} We attach sample weights in different configurations to the origami structure simulating the mass of the solar cells and electronic components. We perform drop tests in each configuration to understand the effect of weight distribution on falling behavior.}
\vskip -0.1in
\label{fig:weight-test}
\end{figure}

Designing a complete origami microflier that can transition in mid-air requires circuit components that can trigger the actuator. Unlike the actuator which is constrained to attachment at the center, the electronics can be placed anywhere on the rigid origami's faces as they do not bend and connecting traces can be routed across the structure. This raises a question of how to optimally distribute the weight on the structure. We simulate our electronics using  25~mg solder balls (gray circles) and 6~mg squares of 125~{\textmu}m thick FR4 (blue squares). We perform a series of experiments to understand the effects of distributing these weights across the origami structure. We tested three different configurations shown in Fig~\ref{fig:weight-test} where A has no weight, B has the weights evenly distributed at the outer edges, C has the weights un-evenly distributed and centered on the faces, and D has an even distribution at the center.  We perform a series of drop tests from a height of approximately 5~m indoors.

We observe first that even weight distribution achieves a greater difference in falling behavior. In particular we note that configurations such as C cause the structure to start flipping about the axis with greater mass in both folding states. In further outdoor experiments we observe that even weight distribution is important for achieving robust stability. We also evaluate attaching a single payload at the end of the tube extending up from the origami surface along the Z-axis and find similar issues of causing differences in fall behavior. 

\begin{figure}
\vskip 0.05in
\centering\includegraphics[width=.95\linewidth]{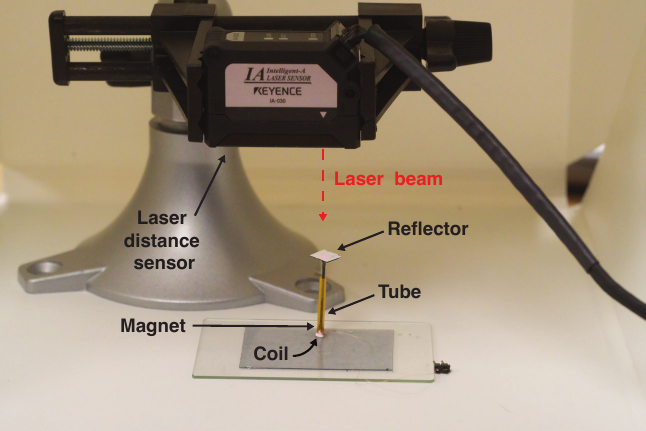}
\vskip -0.05in
\caption{\textbf{Actuator force testing setup.} Test setup used to characterize actuator force. A magnet attached to a carbon fiber rod and flat reflector are placed in a polyimide tube with a soilenoid coil at the base. A laser distance sensor (Keyence IA-030) shining down at the reflector and sampling at 1~kHz was used to measure the distance versus time as the magnet accelerated upward. The actuator was connected to the switch and capacitor circuit used on the origami microflier and probed with an oscilloscope (Tektronix MDO34) to measure the capacitor voltage.}
\vskip -0.1in
\label{fig:actuator-test}
\end{figure}

Next, we tested the effect of evenly distributing the weights at the center or around the edge of the structure. {We observe that evenly distributing the weight of the electronics around the exterior of the structure as in Fig~\ref{fig:weight-test}B maximized the difference in the behavior between the two stable states and minimized the terminal velocity of the structure in the tumbling state. The three and four unit cell designs had the lowest terminal velocities when compared with the five through eight unit cell designs, shown in Fig~\ref{fig:terminal-velocity}.} At a high level, this is likely because in the flat state, the lowest moment of inertia is across one of the axes on the surface of the flat leaf-out structure. However, in the closed state the lowest moment of inertia is now shifted to being along the axis parallel to the tube-coil component. We note that for a freely falling and flipping structure outdoors other aerodynamic effects and lateral winds will affect  the microflier's structure as well.

\begin{figure}
\vskip 0.05in
\centering\includegraphics[width=0.99\linewidth]{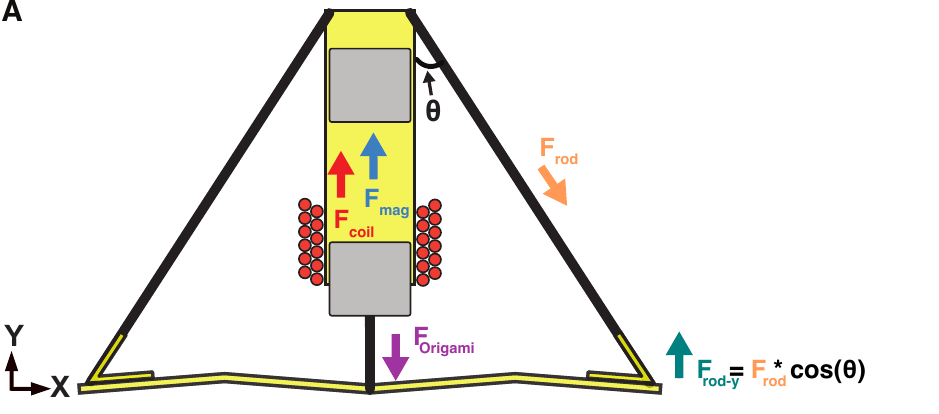}
\vskip -0.1in
\caption{\textbf{Force diagram.} Diagram showing the forces between the actuator and origami structure.}
\vskip -0.2in
\label{fig:force-diagram}
\end{figure}

\subsection*{Microcontroller software and sensor interface}
Our origami microfliers are controlled by a programmable, low-power microcontroller with an integrated Bluetooth radio (Nordic Semiconductor nRF52832). We first describe the design decisions behind our choice of microcontroller followed by details of the embedded software.

{Prior work on wind dispersed devices has demonstrated highly miniaturized devices with electronics weighing less than 30~mg within an approximately 8~mm$^3$ volume~\cite{dandelion-sensors}. Achieving this level of miniaturization however requires certain compromises in functionality such as using a microcontroller without an active radio. These fliers could not receive data transmissions and operated at reduced transmission range and greater hardware complexity such as a full duplex radio at the remote receiver. In this work we incorporate onboard actuation which requires both additional mass and a larger structure to increase power harvesting and drag. With this larger power and size envelope we choose to enable more robust wireless communication and computation functionality by using the nRF52832 Bluetooth chip.}

{Bluetooth is an active radio technology requiring $\sim$15 mW during transmission which necessitates greater power harvesting and energy storage capabilities. Further, it consumes a minimum weight of approximately 50 mg due to the use of a larger chip (3x3 mm compared to 1.65x1.65 mm in~\cite{dandelion-sensors}) as well as a crystal oscillator to provide a frequency reference. The electronics for wireless communication are no longer the limiting factor when scaling up our fliers to support actuation, so we choose to use the nRF52832 Bluetooth chip as our computing platform which comes with upgraded wireless capabilities such as transmitting at bit rates of 1 Mbps and enables receiving data onboard. This onboard receiver enables reception of commands to trigger shape change in mid-air and is a core building block to be able to design networks of these devices. The active transmission enables longer ranges without a full duplex radio and direct communication to ubiquitous Bluetooth receivers integrated into smartphones, laptops etc. The chip also has substantially greater computing power (64 MHz processor ARM Cortex-M4 processor with floating point, 64 KB RAM, 512 KB FLASH) compared to~\cite{dandelion-sensors} (ATtiny20, 8 MHz, 128B RAM, 2KB FLASH).}

\begin{figure}
\vskip 0.05in
\centering\includegraphics[width=0.99\linewidth]{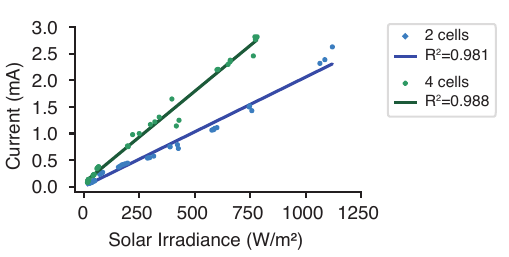}
\vskip -0.1in
\caption{\textbf{Solar irradiance to current.} This plot shows current harvested from two and four solar cells at different light levels. We apply linear regression to obtain a mapping with $R^2 >0.98$ for both ($N \geq 40$ measurements).}
\label{fig:light-to-current}
\end{figure}

Next, we describe the embedded software required to sample data from the onboard sensor and transmit Bluetooth packets. Upon startup, the microcontroller begins by initializing its timers, Bluetooth radio, and I2C interface for reading sensor data. To measure its altitude, our microfliers include an onboard sensor (Bosch BMP384) which measures pressure for mid-air transitions and temperature for environmental monitoring after deployment. The microcontroller uses its I2C interface to send commands and receive data. We note that many other sensors use this standard interface.

{We demonstrate multiple modes of operation. The simplest and lowest power method is to use a timer to transition. In this mode, after the initialization steps described above, the microcontroller immediately enters its sleep mode. Then, upon receiving an interrupt, it wakes up, increments its counter and checks whether it is at the threshold count to transition, and goes back to sleep.} The interrupt for timing can be implemented either using the onboard timer, or by configuring the sensor to provide periodic interrupts. We implement the onboard timer method, but note that the latter solution has the potential for greater power savings by putting the chip into its deepest sleep mode. 

\begin{figure}
\vskip 0.05in
\centering\includegraphics[width=0.75\linewidth]{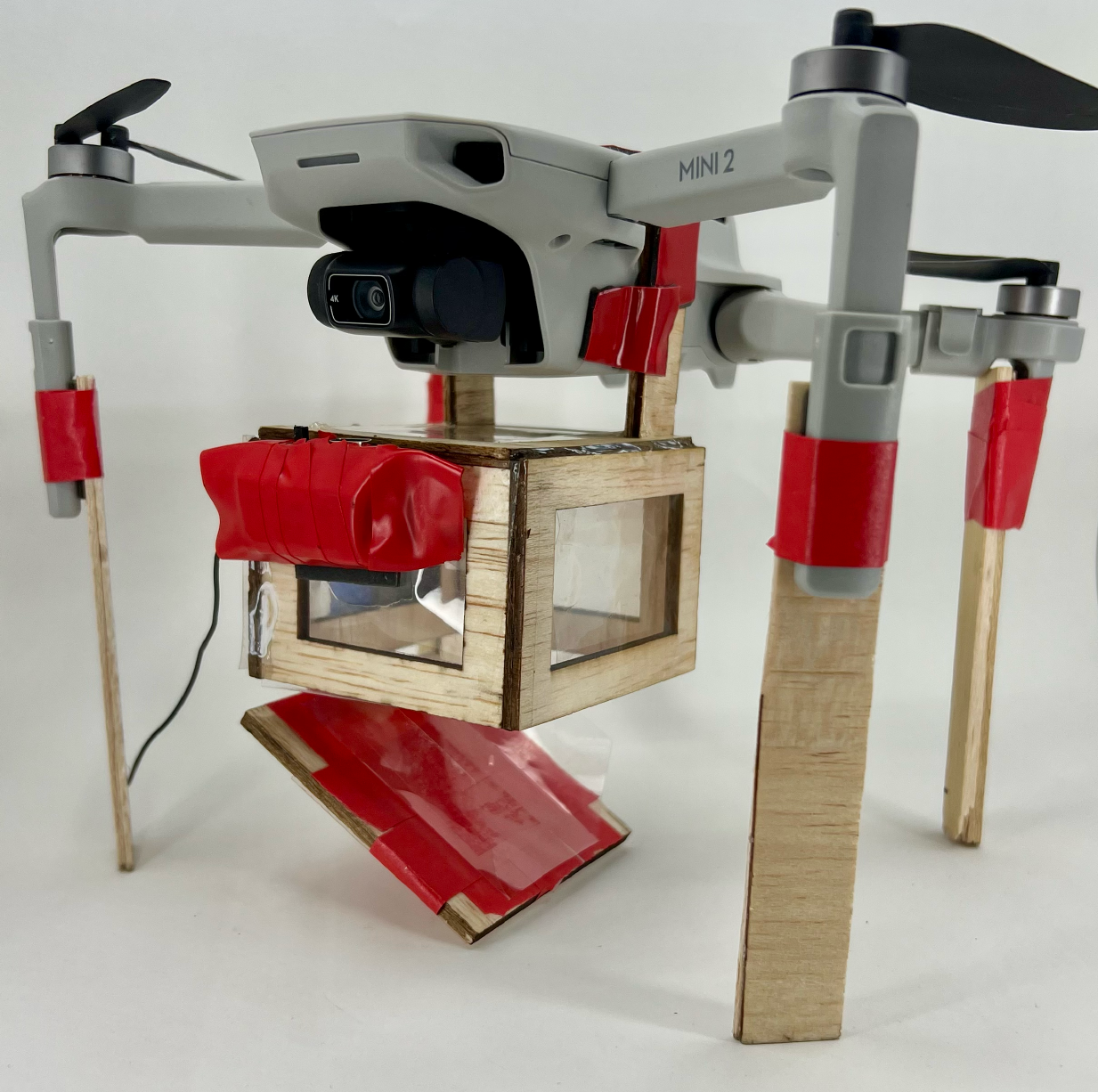}
\vskip -0.1in
\caption{\textbf{Drone and drop mechanism.} Image showing the drone (DJI Mavic Mini 2) and drop mechanism used for deployment experiments.  {The deployment mechanism consists of a balsa wood container with a trap door actuated by a single servo to fit within the payload of the small drone. The servo is connected to a radio receiver that triggers it to open and close. The bottom surface of the trap door is coated with PET to reduce friction and allow the microflier to reliably fall. We note that this design could be further optimized through the use of lightweight high strength materials such as carbon fiber or the use of a larger drone which could carry more complex release mechanisms. We observe that the microflier does not require a fixed position and always descends in the direction of the wind regardless of its initial orientation in the drop mechanism.}}
\label{fig:drone-drop}
\vskip -0.2in
\end{figure}

We also show that during its periodic wake-ups, our flier can also harvest enough power to sample its sensor and transmit data in mid-air. Specifically, we use Bluetooth Beacon or advertising packets which operate in a broadcast mode without requiring an active connection. Fig~7B,C show real time sensor measurements from the microflier in the air as they fall. In these experiments, we program the chips to transmit at a rate of 1~Hz, which can be increased depending on power availability.

\begin{figure}
\vskip 0.05in
\centering\includegraphics[width=0.99\linewidth]{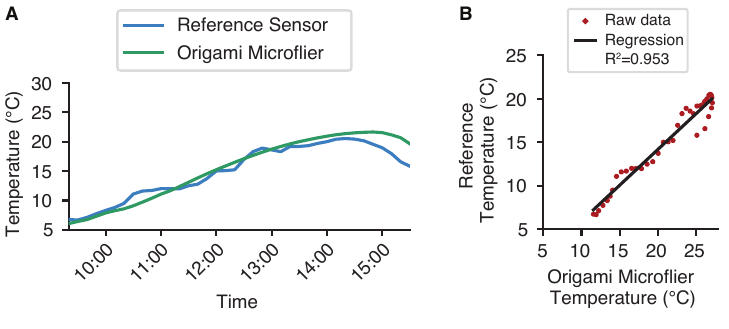}
\vskip -0.15in
\caption{\textbf{Sensor evaluation.} \textbf{A)} Day long comparison of temperature sensor data from our deployed origami microflier and a reference sensor {demonstrating a similar trend after a fixed offset subtraction to compensate for differences in placement and thermal mass of the reference sensor. \textbf{B)} Linear regression showing high correlation between the microflier sensor and the reference with $R^2 >0.95$ ( $N \geq 9000$ 1~Hz measurements).}}
\label{fig:sensor-eval}
\end{figure}

In the second mode, the microflier uses its onboard pressure sensor to trigger a transition at a programmed altitude. The microcontroller again goes to sleep and periodically wakes off of a timer interrupt. In this mode, it sends a command to read the pressure sensor and checks the value against the threshold to determine whether or not to transition. In certain deployment scenarios, this however raises a challenge: if the microcontroller is enabled on the ground and the sensor remains ON while the drone is ascending, it could trigger a false transition. To prevent this, we can add a simple logic check that stores the previous pressure value and checks the delta to make sure the pressure is increasing indicating the sensor is falling. Fig~7B shows the readings from our pressure sensor both when ascending to 50~m and the real time pressure values when dropped. The data show that the resolution of the pressure sensor is sufficient to detect altitude as the maximum altitude matches the 50~m shown by the drone's altimeter. Further, the sensor was programmed to wake up and transmit data at 1~Hz during this experiment, demonstrating that the device can detect changes in altitude even at a low sampling rate. We also observe that although the rate of descent changes due to the wind, the altitude consistently decreases. We note one exception at an altitude of around 10~m. Single point errors like this could also be eliminated by comparing the gradient across two or more points.

\begin{figure}
\vskip 0.05in
\centering\includegraphics[width=1.0\linewidth]{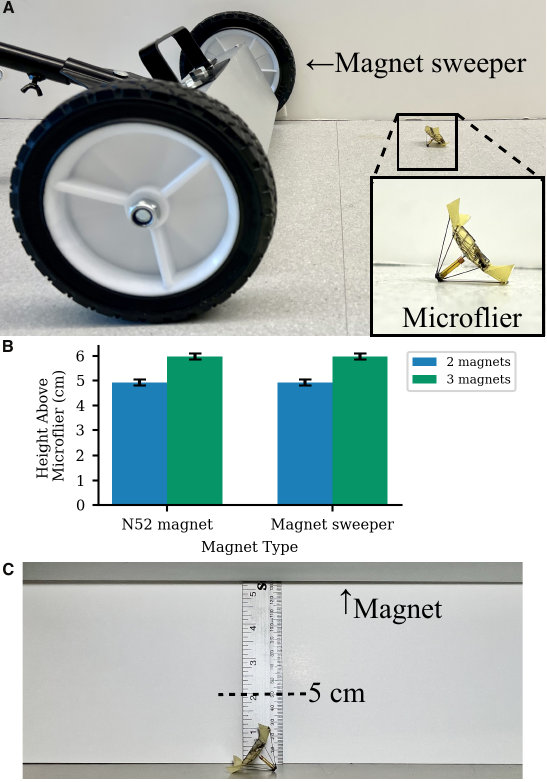}
\vskip -0.1in
\caption{{\textbf{Magnet pick up tests.} \textbf{A)} Demonstration of automated microflier pick up using a magnet sweeper. \textbf{B)} Average height that the microflier is attracted to the pickup system using two different magnets ($N=10, \pm\sigma$). \textbf{C)} Test setup for data gathered in B).}}
\label{fig:magnet-pickup}
\end{figure}

The third mode leverages the onboard Bluetooth radio to transition. We demonstrate that even with limited solar power, it is possible to run a receiver for remote controlled transitions. While transmissions can be easily duty cycled and only require enabling the power hungry Bluetooth radio for a few hundred microseconds, in contrast receiving data requires running the radio for longer. The chip allows a minimum receiving window of 2.5~ms. Moreover, unless the transmitter and receiver are perfectly synchronized (which is difficult to achieve with an intermittent solar power source) the receiver does not know when a packet will arrive. In order to enable packet reception with minimum power, we set the radio to turn on once per second for a short window of 5~ms. To balance the additional power draw, we reduce the rate of Bluetooth sensor transmissions to 0.5~Hz. We use a second nRF52 chip on the ground programmed to send short packets at its maximum rate of over 20~ms to trigger the transmission. Due to the short duty cycle of the receiver, we observe substantial variance of roughly 30~s before the microflier receives the command. While the duty cycle could be increased in brighter sun conditions, in low light conditions we instead use the transmission to set a programmable counter on the device similar to the first mode described above. This allows flexibility and control for such manually triggered deployment scenarios and we use this method to conduct experiments. The ability to receive Bluetooth data can also be used to remotely program an altitude threshold. To reduce the cold-start time of the circuit in-air, we designed the drop mechanism system to be clear. This allows the onboard capacitors to begin charging from the time the microflier is first brought outside, through when the microflier is in the drop container while the drone is flying to the drop location and altitude.

\subsection*{Outdoor sensor evaluation}
In addition to make sure our microflier circuits can cold-start successfully, we also evaluate on-board sensor performance. Fig~\ref{fig:sensor-eval}A shows a comparison of the data from our microflier against a reference sensor (Sparkfun Openlog Artemis) with a fixed offset subtracted. The data shows that the signals are closely correlated. {We observe an offset in the raw data likely due to the fact that our sensor is mounted onto a thin material (12 {\textmu}m) with lower thermal mass than the reference sensor which is attached to a larger PCB ($\approx 1$~mm thick). Additionally, the placement of the reference sensor was constrained by access to power and its exposure to sun and wind could also cause small changes. Fig~\ref{fig:sensor-eval}B shows the correlation along with a linear regression analysis which yields an $R^2 > 0.95$.} The results demonstrate that the sensors successfully perform environmental measurements. 
 {We note however that like with many sensors the results should be calibrated to the specific goals of the application.} 

\begin{figure}
\vskip -0.05in
\centering\includegraphics[width=0.75\linewidth]{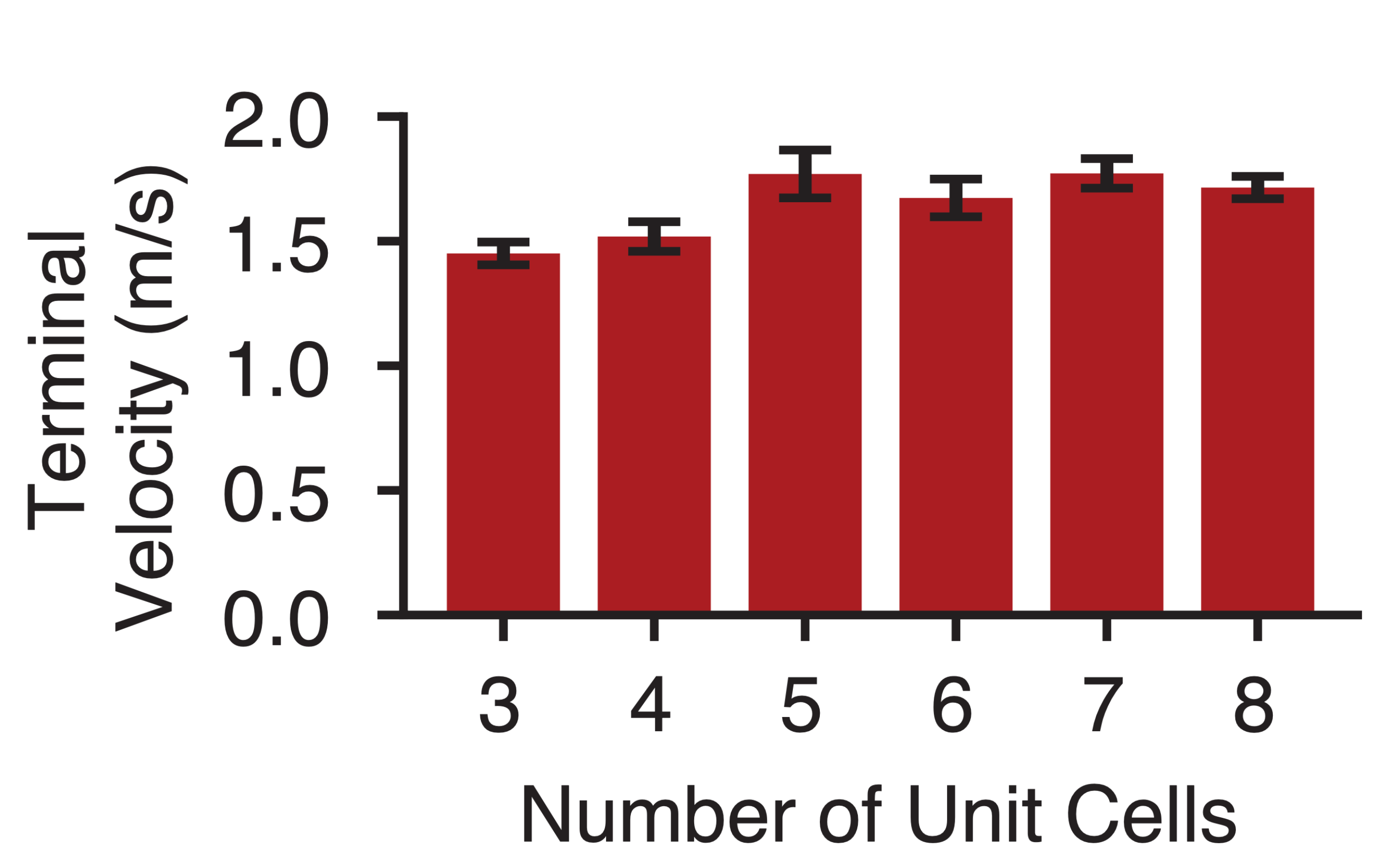}
\vskip -0.15in
\caption{{\textbf{Terminal velocity data.} Average terminal velocity of the leaf-out design with three through eight unit cells when dropped in the tumbling state from a height of 2 m  state ($N \geq 10, \pm\sigma$).}}
\label{fig:terminal-velocity}
\vskip -0.25in
\end{figure}

\subsection*{Robustness and reuse}

\begin{figure*}[t]
\vskip -0.3in
\centering\includegraphics[width=0.95\linewidth]{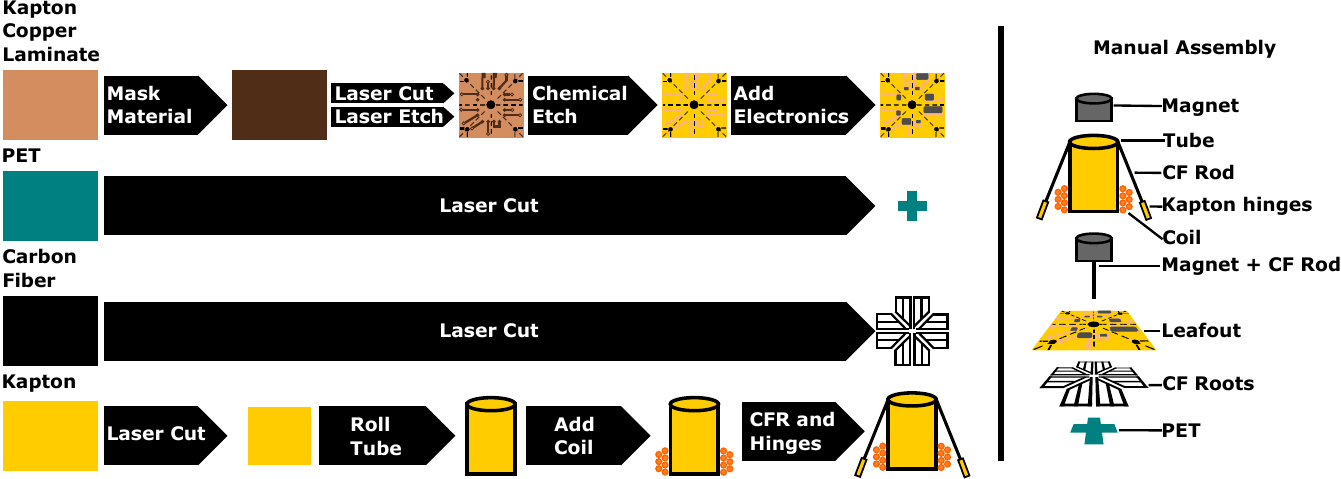}
\vskip -0.1in
\caption{{\textbf{Pictorial flow diagram of fabrication process.} The flow diagram outlines the fabrication steps for each part of the origami microflier and denotes which assembly processes are automated versus the processes that require manual assembly.}}
\label{fig:flow-diagram}
\end{figure*}

\begin{figure*}[h]
\vskip -0.1in
\centering\includegraphics[width=0.7\linewidth]{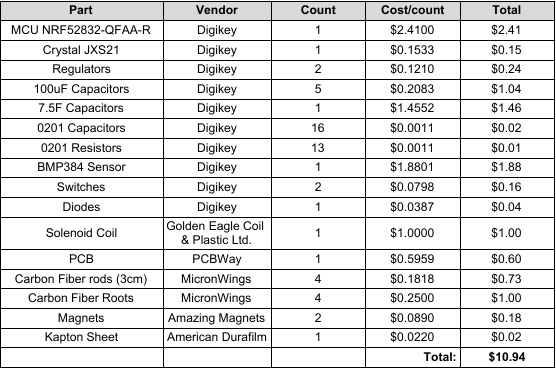}
\vskip -0.1in
\caption{{\textbf{Microflier cost estimate.} The total cost for a single origami microflier is approximately \$10.94. The part cost for components are obtained from the electronics distributor Digikey and are the unit prices for order quantities of 10,000 (prices as of 5/15/2023). We estimate each part from our finalized circuit schematic unit price for order quantities of 1000. The PCB cost is based on a quote for an order quantity of 1000 of our final circuit design from manufacturer PCBWay. Bulk materials such as Kapton films are estimated based on the amount of material used. Custom made solenoid coils were used in the actuation mechanism. This price estimate excludes the cost of solar cells as we are unable to obtain volume price estimates from our solar cell supplier.}}
\label{fig:cost}
\end{figure*}

We performed approximately 50 drop tests from 40~m using a single microflier, demonstrating that our device has the potential for multiple use cycles. Additionally, we observe in our experiments that even the few prototypes that failed mechanically were able to sample and transmit sensor data without issue. This enables devices to be value cycled as static wireless sensors even after mechanical failure. At the end of the microfliers' life cycle we demonstrate how a sweeping magnet could pick up the devices to reduce the environmental pollution from this technology, shown in Fig~\ref{fig:magnet-pickup}A-C.

We also note that our experiments were performed in the fall in Seattle, WA, USA during which our prototypes were exposed to high humidity and moistrure in the form of dew, and light rain during field experiments. Resistance to moisture can be further improved through the use of conformal coatings which are common in the consumer electronics industry for waterproofing circuits. Specific coating products include Microcure DTO (Cytonix) and Teflon AF which the authors have used previously year long outdoor deployments~\cite{eclipse-ipsn}. For example, a wet film thicknesses of 7.5~{\textmu}m  of Microcure DTO can achieve a hydrophobic coating with water contact angles of 100{\textdegree} while weighing 5.6 mg on our microflier.
\section*{Supplementary Movies} 
\noindent\textbf{Movie S1: Origami microflier behavior in wind tunnel.} Our bi-stable origami microfliers exhibit different falling behaviors in their two states over. In the flatter ``tumbling state", they rotate and flip as they fall. In the ``stable state" folded inwards, they exhibit a stable straight downward descent. We compare these behaviors over a vertical wind tunnel first with the base origami structure, followed by our complete microflier with its onboard circuit and actuator. \\

\noindent\textbf{Movie S2: Origami microflier compared to falling leaves.} Our origami microfliers are shown in free fall in their two states compared to leaves with similar behaviors. \\

\noindent\textbf{Movie S3: Origami microflier in lateral wind.} Our origami microflier experiences greater displacement due to lateral wind gusts in its tumbling state than in its stable state. The video shows a fan blowing air to the right as the microflier is dropped in each of its states. In its stable state, the microflier moves to the right substantially when it encounters the wind whereas in its stable state the wind has minimal effect. \\

\noindent\textbf{Movie S4: Origami microflier manual transition.} The video shows a diagram of our microflier's origami structure and manual folding to demonstrate how it transitions between its two stable states. \\

\noindent{\textbf{Movie S5: Origami microflier false transition tests.} We tested the robustness of our microflier's bi-stability by hanging one of our prototypes from a thin Kevlar thread vertically above a fan. We recorded videos of the microflier in two different wind speeds, and tracked the fan's wind speed from an anemometer that actively output the wind speed data to a phone that is visible in the video. The video shows that the microflier does not transition even at wind speeds of 5~m/s. We observe the same result for N=10 trials averaging 7~s in duration.} \\

\noindent\textbf{Movie S6: Electronically actuated origami transition.} 
The video shows a tethered demonstration of how our light-weight, low-power solenoid actuator can produce the required force and bending motion to transition the origami structure from its tumbling state to its stable state. Our actuator produces substantial force causing the structure to briefly lift off the table. \\

\noindent\textbf{Movie S7: Untethered solar-powered transition.} Demonstration of a fully assembled and untethered origami microflier transitioning between states using harvested solar power. The onboard, battery-free circuit is able to harvest sufficient solar energy to start itself up and begin  listening for Bluetooth signals. Upon receiving a packet with the command to transition, it triggers the actuator. The onboard microcontroller can be  programmed to transition from a remote signal, onboard timer, or pressure sensor at a particular altitude. \\

\noindent\textbf{Movie S8: Mid-air solar-powered transition from low altitude.} Origami microflier transitioning in mid-air using harvested solar power when dropped from a ladder at a height of approximately 4 m. The video begins by showing the full trajectory and then proceeds to show a zoomed in, slow motion view pausing to highlight the transition point. We note this version of the microflier is an early prototype with thin wires for debugging and testing visible in the video. \\

\noindent\textbf{Movie S9: Mid-air solar-powered transition from drone deployment.} Origami microflier transitioning in mid-air using harvested solar power after being dropped from a drone at an altitude of 20 m. The origami microflier starts falling in its tumbling state before transitioning mid-air to the stable state. The video shows the full descent trajectory, then proceeds to show a zoomed in and slow motion view to highlight the transition point.  \\ 

\noindent\textbf{Movie S10: Evaluating dispersal distance.} We evaluate wind dispersal distance by dropping our origami microfliers from drones at multiple altitudes and in various wind conditions. Recording a drone deployment video from 40~m in higher wind speeds requires running after the microflier as it travels 86~m, resulting in shaking of the camera and difficulty keeping the microflier in the frame.

\end{document}